\documentclass[sigconf, screen=true]{acmart}
\usepackage{booktabs} % For formal tables
\usepackage{lipsum}
\usepackage{multirow}
\usepackage{enumitem}
\usepackage[normalem]{ulem}
\useunder{\uline}{\ul}{}

\setcopyright{rightsretained}
\usepackage{color}
\definecolor{mygreen}{rgb}{0.9,0.65,0.8}
\definecolor{mylightorange}{rgb}{0.9,0.72,0.55}
\definecolor{mycyan}{rgb}{0.7,0.9,1}
\definecolor{mygray}{rgb}{0.92,0.92,0.92}

\newcommand{\ours}{TailCalibration}

\newcommand{\mcD}{\mathcal{D}}
\newcommand{\mcB}{\mathcal{B}}
\newcommand{\mcF}{\mathcal{F}}
\newcommand{\mcN}{\mathcal{N}}
\newcommand{\xvec}{\mathbf{x}}
\newcommand{\yvec}{\mathbf{y}}
\newcommand{\zvec}{\mathbf{z}}
\newcommand{\svec}{\mathbf{s}}
\newcommand{\pvec}{\mathbf{p}}
\newcommand{\wvec}{\mathbf{w}}
\newcommand{\ztilde}{\tilde{\mathbf{z}}}
\newcommand{\Wmatrix}{\mathbf{W}}
\newcommand{\real}{\mathbb{R}}

\usepackage{xspace}
\makeatletter
\DeclareRobustCommand\onedot{\futurelet\@let@token\@onedot}
\def\@onedot{\ifx\@let@token.\else.\null\fi\xspace}

\def\eg{\emph{e.g}\onedot} 
\def\ie{\emph{i.e}\onedot}

\def\etal{\emph{et al}\onedot}
\makeatother

\acmDOI{10.1145/3490035.3490300}

\acmArticle{43}

\acmISBN{978-1-4503-7596-2}

\acmConference[ICVGIP'21]{12th Indian Conference on Computer Vision, Graphics and Image Processing}{December 2021}{Jodhpur, India}
\acmYear{2021}
\copyrightyear{2021}

\acmPrice{15.00}

\begin{document}
\title{Feature Generation for Long-tail Classification}
% \titlenote{Produces the permission block, and copyright information}

% \author{Rahul Vigneswaran}
% \affiliation{\institution{Indian Institute of Technology, Hyderabad}\country{India}}
% \author{Marc T. Law}
% \affiliation{\institution{Nvidia}\country{Canada}}
% \author{Vineeth N. Balasubramanian}
% \affiliation{\institution{Indian Institute of Technology, Hyderabad}\country{India}}
% \author{Makarand Tapaswi}
% \affiliation{\institution{IIIT Hyderabad}\country{India}}

\author{Rahul Vigneswaran$^1$, Marc T. Law$^2$, Vineeth N. Balasubramanian$^1$, Makarand Tapaswi$^3$}
\affiliation{%
$^1$Indian Institute of Technology, Hyderabad \quad
$^2$NVIDIA \quad
$^3$IIIT Hyderabad
\country{India}
}

% \author{Submission Id 160}
% \affiliation{%
%   \institution{XYZ}
%   \streetaddress{XYZ}
%   \city{XYZ}
%   \state{XYZ}
%   \country{XYZ}
%   \postcode{000000}
% }

\renewcommand{\shortauthors}{R. Vigneswaran, et al.}

\begin{abstract}
The visual world naturally exhibits an imbalance in the number of object or scene instances resulting in a \emph{long-tailed distribution}.
This imbalance poses significant challenges for classification models based on deep learning.
Oversampling instances of the tail classes attempts to solve this imbalance. However, the limited visual diversity results in a network with poor representation ability.
A simple counter to this is decoupling the representation and classifier networks and using oversampling only to train the classifier.

In this paper, instead of repeatedly re-sampling the same image (and thereby features), we explore a direction that attempts to generate meaningful features by estimating the tail category's distribution.
Inspired by ideas from recent work on few-shot learning~\cite{yang2021freelunch}, we create calibrated distributions to sample additional features that are subsequently used to train the classifier.
Through several experiments on the CIFAR-100-LT (long-tail) dataset with varying imbalance factors and on mini-ImageNet-LT (long-tail), we show the efficacy of our approach and establish a new state-of-the-art.
We also present a qualitative analysis of generated features using t-SNE visualizations and analyze the nearest neighbors used to calibrate the tail class distributions.
Our code is available at \url{https://github.com/rahulvigneswaran/TailCalibX}.
\end{abstract}




\maketitle

\section{Introduction}
\label{sec:intro}

\begin{figure}[t]
\centering
\includegraphics[trim={8cm 3cm 6cm 0},clip, width=\linewidth]{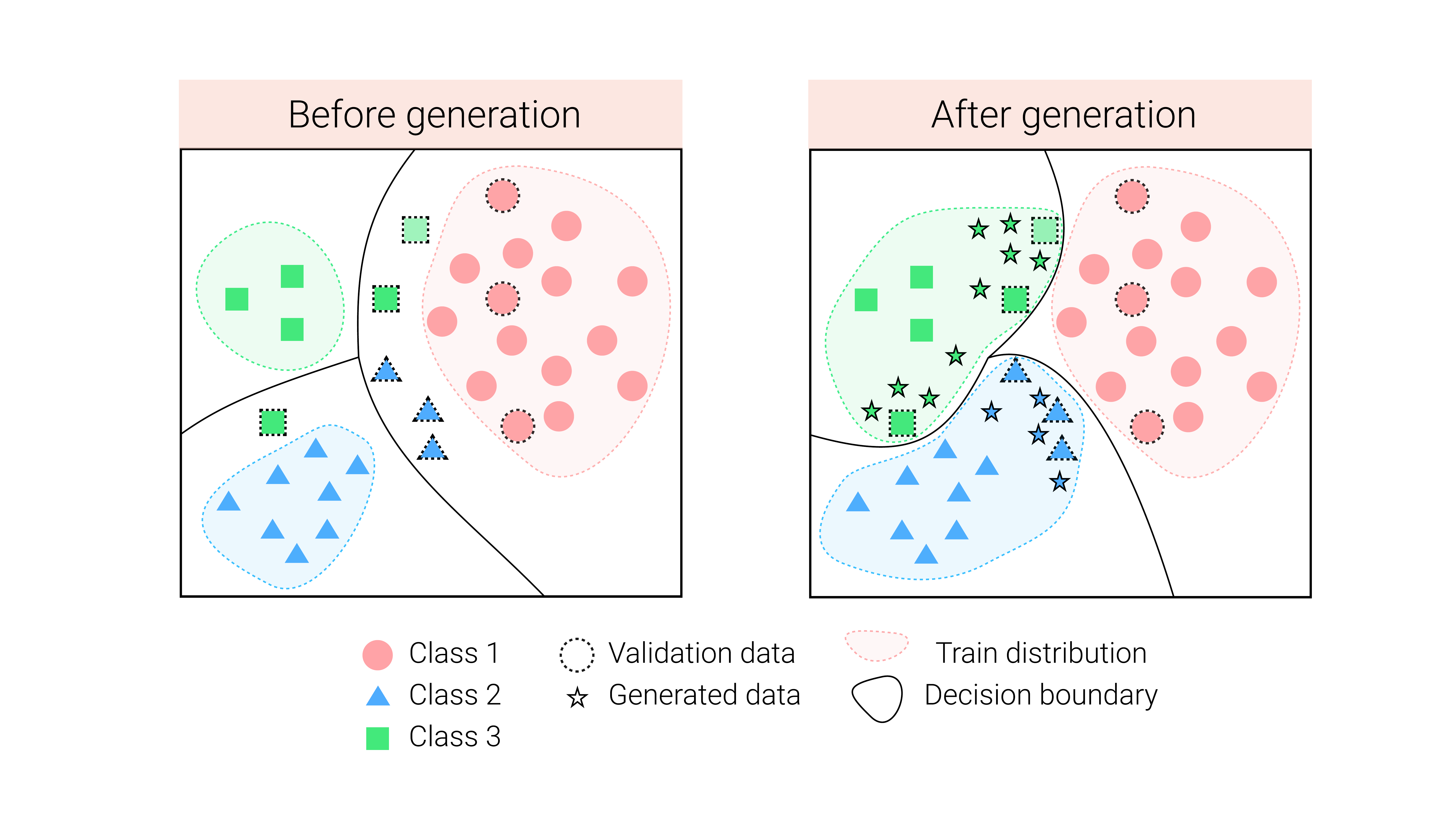}
\vspace{-3mm}
\caption{Long-tail distributions consist of few categories with many samples (head of the long tail, in red) and many categories with few to very-few samples (green, blue).
Due to the larger diversity of the head classes, decision boundaries are often favourable for the head class, while being error-prone for the tail classes.
The goal of our work is to generate meaningful additional features for the tail classes so that a balanced training set is created for the classifier. 
To this end, we estimate the distribution of the tail classes based on individualized instances and sample additional features through this \emph{calibrated tail distribution}.}
\label{fig:teaser}
\vspace{-2mm}
\end{figure}

%-- balanced datasets are mainstay
Modern machine learning is driven by large scale datasets, employed in both supervised~\cite{russakovsky2015imagenet,parkhi2015vggface,gemmeke2017audioset} and self-supervised~\cite{sim2sg,miech2019howto100m,sun2017jft,sharma2018conceptual} scenarios.
However, creation of these labeled datasets is a challenging and costly affair~\cite{acuna2021fdomainadversarial} and may also lead to unforeseen biases~\cite{tsipras2020imagenet}.
Often, these datasets are created by querying images through a search engine~\cite{russakovsky2015imagenet,parkhi2015vggface}, followed by post-processing and ``cleaning'' to ensure a \emph{balanced} distribution that has an (approximately) equal number of instances for each category.

%-- but the world is long-tail
However, the world we live in is naturally \emph{long-tail}, and like the language modality, even visual data follows the Zipf's law~\cite{zipf1949,vanhorn2018inaturalist}.
This is easily illustrated through examples we encounter in our daily lives: people living in a city are more likely to see multiple instances and a large diversity of cars than elephants for transportation, and tables and chairs than tree stumps as furniture.
This natural distribution of categories is reflected in the datasets that are collected through community efforts, \eg~iNaturalist~\cite{vanhorn2018inaturalist} that features a large image collection of biodiversity; or annotations for a collection of randomly sampled raw data -- action labels from the Atomic Visual Actions dataset~\cite{gu2018ava}; or interaction and relationship labels from movie datasets~\cite{vicol2018moviegraphs}.

%-- several techniques in the recent years
Modern deep learning methods perform well on balanced distributions and have even shown super-human performance for some datasets and tasks in diverse domains including image classification~\cite{he2015superhumanimagenet} and language understanding~\cite{wang2019glue}.
This has led to a steadily growing interest in adapting methods to work well with few training samples -- \emph{few-shot learning}~\cite{Cao2020A,wang2019fslsurvey,dhillon2020fslbaseline,mangla2020manifoldfsl,8683393} or naturally occurring \emph{long-tail distributions}~\cite{kang2019decoupling,lin2017focal,zhou2020bbn,tang2020gbm,iscen2021cbd}.
Interestingly, it is observed that popular techniques used in shallow learning (\eg~SVMs) such as loss re-weighting or balanced sampling~\cite{japkowicz2000sampling} do not perform well when they are applied with deep neural networks, as they may hurt the representation learning~\cite{kang2019decoupling,zhou2020bbn}.
Instead, decoupling the feature learning (\ie~CNN backbone) from the classifier (\ie~final linear layers) is found to be necessary and useful~\cite{kang2019decoupling}.

%-- problems with those techniques, feature generation as an alternative
In this work, we are interested in answering the following question.
Can we learn to generate feature representations for the impoverished tail classes, instead of requiring re-weighting (\eg~\cite{zhou2020bbn}) or multi-network distillation (\eg~\cite{iscen2021cbd}) strategies?
Inspired by the recent success of \emph{distribution calibration} on few-shot learning~\cite{yang2021freelunch}, we estimate and calibrate the feature distribution of tail classes to sample additional features that are in turn used to train a classifier - we term this as \emph{\ours{}} (see Fig.~\ref{fig:teaser}).
Note that our work is different from~\cite{chawla2002smote,kim2020m2m,dablain2021deepsmote} that attempt to reconstruct or generate images for the tail class; or from~\cite{zhang2017mixup,verma2019manifoldmixup} that use convex combinations of input samples or features \emph{and} their labels.
\ours{} is agnostic to the training approach of the backbone, and therefore, adopting better classification setups (\eg~\emph{CosineCE}~\cite{luo2018cosinece}) or using distillation (\eg~CBD~\cite{iscen2021cbd}) can further improve overall performance.

%-- summarize contributions
Our contribution can be summarized as follows:
We explore feature generation as a means to address the challenges of long-tail classification, and show that \ours{}, an adapted version of~\cite{yang2021freelunch}, is an effective strategy to generate meaningful additional features (Sec.~\ref{sec:method}).
We empirically validate our approach through multiple ablation studies on the CIFAR-100-LT achieving the state-of-the-art performance (Sec.~\ref{sec:experiments}).
Our analysis also holds for a long-tailed version of a similarly sized dataset called  \emph{mini-ImageNet-LT}~\cite{vinyals2016matching} that had been introduced for few-shot learning.
We also present a qualitative analysis of generated features using t-SNE visualizations and analyze the category-level nearest neighbors used to calibrate the tail class distributions.

\section{Related Work}
\label{sec:relatedwork}

We discuss and contrast related work on long-tail classification with our proposed approach.
While imbalanced learning has had a long history, we focus primarily on modern techniques applied to deep learning.
In general, we can group related works in 4 broad categories:
(i) modifications to the loss function or re-weighting of samples;
(ii) decoupling the learning process of the representation and classifier;
(iii) using distillation or multiple experts; and
(iv) perhaps closest to our work, ideas related to sample or feature generation.

\subsection{Re-weighting or Adapting the Loss Function}
\label{subsec:rw-loss}

Class re-weighting or balanced sampling are popular approaches to deal with long-tail datasets especially in shallow architectures~\cite{japkowicz2000sampling}.
However, with deep learning, as the same loss function trains both the representation network and the classifier, the above ideas are not directly applicable.

The cross-entropy (CE) loss is modified to down-weight \emph{easier} examples while focusing on harder ones~\cite{lin2017focal}, or to include the effective number of training samples via a re-weighting term~\cite{cui2019classbalanced}.
As an alternative to CE, there has also been work on deriving a class-specific margin parameter to be inversely proportional to the fourth root of number of samples for a margin-based loss~\cite{cao2019ldam}.
Long-tail classification and label corruption are also attempted simultaneously through meta-learning networks for re-weighting instances~\cite{ren2018metareweight,shu2019metaweightnet}.
Recently, a simple \emph{logit adjustment} that can be applied post-hoc to trained models~\cite{menon2021logitadjustment} is proposed as a means to unify many of the above approaches.

A few different perspectives applied to long-tail classification include viewing it as a domain adaptation task (target shift)~\cite{jamal2020domainadaptation, hong2021disentangling}, or as a causal framework where stochastic gradient descent's momentum term is treated as a confounder~\cite{tang2020gbm}.

Our work does not belong to this category and can be thought of as orthogonal to any of the techniques above.

\subsection{Decoupling Representation and Classifier}
\label{subsec:rw-decoupling}

One of the key challenges of deep learning on long-tail datasets is exposed by~\cite{kang2019decoupling,zhou2020bbn}.
Both works demonstrate that
(i) representation learning suffers when applying balancing or re-weighting techniques on end-to-end training paradigms, and
(ii) classifier performance of tail classes is adversely impacted when using standard instance sampling.
As remedies, a two stage approach is suggested.
In the first stage, usual end-to-end training is performed using instance sampling.
In the second stage, the classifier is decoupled from the network and is trained with balanced sampling~\cite{kang2019decoupling}.
While this two-stage strategy improves the performance, \cite{zhong2021improving}~discovers that this may lead to domain shift between the representation and the classifier parts when seen from the perspective of transfer learning and proposes MiSLAS - a batch normalization based trick to resolve this.
On similar lines, a dual network with shared backbone is proposed by~\cite{zhou2020bbn} with an adaptive trade-off parameter that balances the two objectives.

Our work is related to the above - we also decouple the training of the backbone from the classifier.
The key difference is that instead of balanced sampling, we attempt to generate meaningful features for the tail classes that are subsequently used to train the classifier.

\subsection{Distilling Information across Networks}
\label{subsec:rw-distill}

An alternative approach to long-tail classification is using (multiple) teachers to distill information into a student network.
One idea here involves learning separate classifiers for sets of categories such that the segregated imbalance factor is reduced before combining them~\cite{xiang2020lfme}.
Along those lines, RIDE~\cite{wang2020long} trains multiple experts with shared earlier layers and routes the inference through a subset of trained experts to improve tail class performance. These subset of experts can be self-distilled from the a larger subset of experts for further improvement.
An alternative idea considers an ensemble of teachers trained on instance sampling that are used to uphold the quality of the representation network while training the student on a balanced sampling schedule~\cite{iscen2021cbd}.

Our strategy is orthogonal to this approach and we show that applying \ours{} on the distilled model from~\cite{iscen2021cbd} further improves performance.

\subsection{Instance or Feature Generation}
\label{subsec:rw-gen}

The final set of approaches address the challenges of long-tail classification through generation or augmentation of samples or features.
Among \emph{sample generation}, SMOTE~\cite{chawla2002smote} is an old but popular choice that generates new instances through interpolation between samples of the same (often, the tail) class.
Parallel to our work, SMOTE has been extended to deep learning where an encoder-decoder framework is used to generate additional images, DeepSMOTE~\cite{dablain2021deepsmote}.
Similar to SMOTE, \emph{mixup} is a general supervised learning paradigm that allows models to learn from convex combinations of not only input samples, but also their labels~\cite{zhang2017mixup}.
Specifically designed for visual long-tail classification, M2m translation transfers the diversity of samples from the head class to the tail through gradient updates, resulting in minor visual transformations that lead to significant impact on classification scores~\cite{kim2020m2m}.
Similar to M2m, \cite{chu2020feature}~explicitly models class-specific and class-generic features which are then mixed together to transfer the sample diversity from the head class to the tail.

While mixup is applied on the input space, manifold mixup is applied on the semantic (feature) space~\cite{verma2019manifoldmixup} and is effective at generating convex feature combinations.
Specific data augmentation strategies relevant for long-tail classification transfer the implicit knowledge of features from the head classes that exhibit a high variance, to the tail classes that lack intra-class diversity~\cite{liu2020headtailvariance}.
This is done by constructing a feature cloud around the existing tail data points to match the head class variance.
A recent work MODALS~\cite{cheung2021modals} suggests multiple modality agnostic feature augmentation strategies based on hard example interpolation, extrapolation, and Gaussian noise around the samples.

Our work is related to these ideas, but differs in the actual method for feature generation.
By calibrating distributions we expect to learn from the diversity of head classes and generate features for tail classes to balance the inputs seen by the classifier.

\section{Method}
\label{sec:method}

\begin{figure*}[t]
\centering
\includegraphics[trim={2cm 18cm 0 10cm},clip,width=\linewidth]{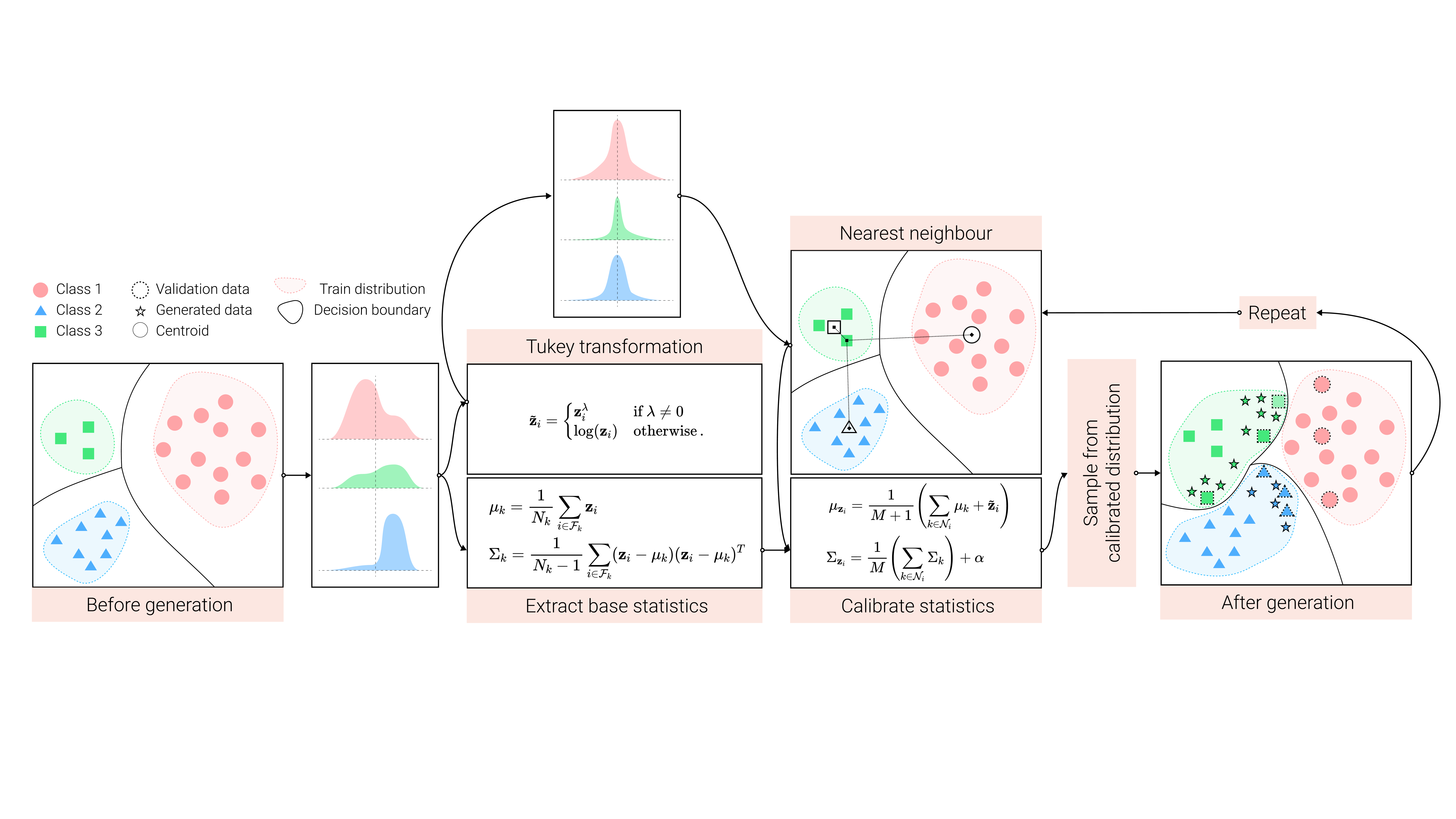}
\caption{We visualize the process of feature generation.
We first estimate the class statistics, and apply the Tukey transformation to convert the data into a distribution closer to a Gaussian.
We find the nearest neighbor categories for each instance, calibrate the tail class distribution, and sample from this to generate additional features.
The process is repeated until all categories have balanced number of features, which are used as training data for the classifier.}
\label{fig:method}
\end{figure*}

We present details of our proposed method, \ours{}, that generates additional features for the tail classes.
We start by formulating the long-tail classification task through some notation and a discussion of multiple strategies to train the backbone network.

\subsection{Problem Definition}
We address $K$ class supervised classification on a training dataset $\mcD = \{(\xvec_i, c_i)\}_{i=1}^N$ with $N$ samples.
$\xvec_i$ represents the $i^{\mathrm{th}}$ image and $\yvec_i$ is the one-hot encoding of the category label $c_i \in \{1, \ldots, K\}$.
For simplicity, we denote the set of samples belonging to category $k$ by $\mcD_k$, and denote its cardinality by $N_k = |\mcD_k|$.
A long-tail setup can be defined by ordering the number of samples per category, \ie~$N_1 \geq N_2 \geq \ldots \geq N_K$ (without loss of generality), such that $\sum_k N_k = N$.
The \emph{imbalance factor} of the dataset is indicated as the ratio of samples in the head to tail class, $N_1 / N_K$, where a higher imbalance often translates to worse performance on the tail categories.

We train a network $\Phi$ consisting of two components:
(i) a \emph{backbone} or representation network (CNN for images) that translates an image to a feature representation $f (\xvec_i) = \zvec_i$ where $\zvec_i \in \real^D$, and
(ii) a \emph{classifier} $\Wmatrix \in \real^{K \times D}$ that predicts the category specific scores (logits) $g (\zvec_i) = \svec_i$.
When not specified otherwise, we assume $\svec_i = g(\zvec_i) = \Wmatrix \zvec_i$.
We ignore writing the bias term of the linear layer for brevity.

\subsection{Training the Backbone}
\label{subsec:backbone}

Our approach on generating features is independent from the manner in which the representation network is trained.
Below, we present three ways of obtaining a trained backbone network $f$.

\paragraph{CE}
We consider a batch-wise training approach with instance sampling, where each sample $\xvec_i$ has equal probability of being considered as part of a mini-batch $\mcB$.
Specifically, the parameters of the backbone $f$ and the classifier $g$ are trained end-to-end using the cross-entropy (CE) loss:
\begin{equation}
\label{eq:ce}
L_{CE} = - \sum_i \yvec_i \log(\pvec_i) \, ,
\end{equation}
where $\pvec_i$ is a vector of probabilities obtained by transforming the logits $\svec_i$ through the softmax operation.

\paragraph{CosineCE}
As an alternative to the standard dot-product classifier, a cosine normalization based classifier is proposed to reduce the variance of the output neuron~\cite{luo2018cosinece}.
The key difference lies in the computation of the dot product -- the feature vector $\zvec_i$ and each row (category $k$) of the classifier, $\wvec_k$, are $\ell_2$-normalized prior to computing the dot product.
The logit score for sample $\xvec_i$ and category $k$ is computed as follows:
\begin{equation}
\label{eq:cosinece_logit}
s^{\mathrm{CosCE}}_{ik} = \gamma\frac{\zvec_i^{\top} \wvec_k}{\| \zvec_i \|_2 \| \wvec_k \|_2} \,  ~~~~~~~~~~~ \text{ where } \zvec_i = f (\xvec_i),
\end{equation}
where $\gamma \in \mathbb{R}_+$ is a positive normalization constant that ensures that $s^{\mathrm{CosCE}}_{ik} \in [-\gamma, \gamma]$.
We treat $\gamma$ as a learnable parameter and use the \emph{softplus} operator to ensure that it remains positive. We recall that the operator is defined as  $\forall x \in \mathbb{R}$, \emph{softplus}$(x) = \log(1 + e^{x})$. 
% We found this to perform slightly better than ReLU \ML{this = softplus? I don't think it is necessary to mention it}.
The loss function used to train the model remains unchanged from the above Eq.~\eqref{eq:ce}.

This formulation has also been adopted by recent works on few-shot learning~\cite{gidaris2018cosinece} and long-tail classification~\cite{iscen2021cbd}.
In particular, the cosine similarity based score computation embeds all samples close to a single point on the hypersphere (\ie a $(D-1)$-sphere embedded in $\mathbb{R}^D$) indicated by the classifier vector.
We believe that this makes the distribution of features on the hypersphere closely resemble a Gaussian distribution or a von Mises-Fisher distribution  \cite{banerjee2005clustering}. 
We will show the benefits of using CosineCE through our ablation studies.

\paragraph{Class-balanced Distillation (CBD)}
Parallel to our work, Iscen~\etal~\cite{iscen2021cbd} show the use of teacher-student distillation with the goal of improving both the backbone representation and the classifier.
We briefly summarize their approach.
First, a teacher network $\Phi_T: (f_T, g_T)$ is trained using standard instance sampling and the CosineCE classifier described above.
Then, a student network $\Phi_S: (f_S, g_S)$ is trained with balanced sampling along with distillation, \ie~the intermediate feature representation $f_S(\xvec_i)$ is constrained to be similar to $f_T(\xvec_i)$, while harnessing the benefits of balanced sampling for the classifier $g_S$.
We decouple the backbone network trained through a combination of the classification and distillation losses, and show that \ours{} can use these improvements to obtain additional performance gains.

\subsection{\ours{}}
\label{subsec:tailcalib}

The primary objective of our work is to generate additional features $\zvec^*$ such that they can be used to \emph{balance} the data that is used to train the classifier.
\emph{Distribution calibration} (DC) was proposed recently in the context of few-shot learning: samples are generated for the few-shot classes by relying on statistics of the base classes~\cite{yang2021freelunch}.
We investigate the applicability of DC in the context of long-tail classification, and present the generation process below.

Our approach can be simplified into three steps, also illustrated in Fig.~\ref{fig:method}.
(i) We first estimate the distribution of each category based on a Gaussian assumption;
(ii) Categorical neighbors of each instance in the tail classes are used to create a calibrated distribution; and
(iii) We sample several new features from these distributions to balance the training data seen by a classifier.

All operations below are in the feature space.
Given a trained backbone (discussed in Sec.~\ref{subsec:backbone}), we first precompute feature representations for the entire dataset.
These features of true samples are denoted by $\mcF = \{\zvec_i\}_{i=1}^{N}$.
$\mcF_k$ denotes features of images corresponding to the subset $\mcD_k$, or belonging to the category $k$.

Due to the inherent randomness in sampling from a distribution, we can generate features multiple times and use them as a \emph{fresh} dataset from which the classifier can learn something new in each epoch.

\paragraph{Class statistics}
We start by computing the statistics for each category, assuming that the feature distribution is Gaussian.
Note that this assumption may be particularly reasonable when using a backbone trained with the CosineCE loss function.
% as indicated above.\todo{What do you mean by "as indicated above"? Where is it? - fixed}
\begin{eqnarray}
\mu_k &=& \frac{1}{N_k} \sum_{i \in \mcF_k} \zvec_i \, , \mathrm{and} \\
\Sigma_k &=& \frac{1}{N_k - 1} \sum_{i \in \mcF_k} (\zvec_i - \mu_k)(\zvec_i - \mu_k)^T \, ,
\end{eqnarray}
where $\mu_k \in \real^D$ and $\Sigma_k \in \real^{D\times D}$ denote the mean and full covariance of the Gaussian distribution for category $k$.

\paragraph{Tukey's Ladder of Powers transformation}
Sometimes referred to as the Bulging rule, this transformation helps change the shape of a skewed distribution so that it becomes closer to a Normal distribution~\cite{tukey1977eda}.
In particular, it is applied to each dimension of the feature as follows:
\begin{equation}
\ztilde_i =
    \begin{cases}
      \zvec_i^\lambda & \text{if $\lambda \neq 0$}\\
      \log(\zvec_i) & \text{otherwise} \, .
    \end{cases}
\end{equation}
where $\lambda > 0$ is a hyperparameter of our model and $\zvec_i^\lambda$ raises each element of $\zvec_i$ to the power $\lambda$. 
In fact, $\lambda \simeq 1$ is found to perform well for most of our experiments.

\paragraph{Calibration and generation}
For each class, we sample $N_1 - N_k$ additional features, such that the resulting feature dataset is completely balanced and all classes have $N_1$ instances.
Sampling is performed based on an instance specific calibrated distribution.
Specifically, each $\zvec_{ik}$ ($i^\mathrm{th}$ feature from category $k$) is responsible for generating $\left[N_1/N_k - 1\right]_+$ features where $[x]_+$ rounds $x$ to the nearest integer greater than or equal to $x$. 
As $N_1 / N_k$ may be a fraction, we randomly choose an appropriate number of $\zvec_{ik}$ so as to obtain a total of $N_1$ samples.

Next, we compute the distances between class means and the selected feature.
For each feature $\zvec_i$, we first compute the distance between the instance and category means as
\begin{equation}
\forall k, \, d_{ik} = \| \ztilde_i - \mu_k \|_2.
\end{equation}
We identify the set of $M$ category indices that are neighbors $\mcN_i$ with smallest distance $d_{ik}$.
Note that the distance computation uses the Tukey transformed feature $\ztilde_i$.

The calibrated distribution is obtained as
\begin{eqnarray}
\mu_{\zvec_i} &=& \frac{1}{M+1} \left( \sum_{k \in \mcN_i} \mu_k + \ztilde_i \right) \\
\Sigma_{\zvec_i} &=& \frac{1}{M} \left( \sum_{k \in \mcN_i} \Sigma_k \right) + \alpha \, ,
\end{eqnarray}
where $\alpha$ is an optional constant hyper-parameter to increase the spread of the calibrated distribution.
We found that $\alpha = 0$ works reasonably well for multiple experiments.

\paragraph{Sampling}
We initialize a multi-variate Gaussian distribution\footnote{using PyTorch's \texttt{MultivariateNormal} distribution.} for each instance $\zvec_i$, and generate new features with the same associated class label as $c_i$.
We denote the generated features for category $k$ as $\mcF_k^*$, and together, $|\mcF_k \cup \mcF_k^*| = N_1$.
This combined set of features is generated for all categories and used to train the classifier $g$.

\paragraph{$\ell_2$-normalized calibration}
For the CosineCE backbone, we $\ell_2$-normalize the features $\zvec_i$ before feeding them to the \ours{} pipeline, both for gathering statistics and Tukey transformation.

% \begin{enumerate}
%     \item CosineCE Training
%     \begin{enumerate}
%         \item Initialize
%         \item Train with instance sampler (Imbalanced data)
%         \item Train for 150 epochs (CIFAR100), 111 epochs (ImageNet)
%     \end{enumerate}
%     \item FreeLunch
%     \begin{enumerate}
%         \item Use the CNN feature extractor to obtain all the representations of train and validation data.
%         \item Collect statistics: Calculate class-wise mean and co-variance of all the training samples.
%         \item Apply Tukey transform on train and validation data
%         \item Balancing
%         \begin{enumerate}
%             \item Select a random sample from the class of interest
%             \item Find K nearest neighbour
%             \begin{enumerate}
%                 \item Find l2 distance between the chosen sample and all the collected means.
%                 \item Now choose the classes of means which are top K closest to the sample at hand
%             \end{enumerate}
%             \item Calibration: Average the statistics (mean and co-variance) of these selected K nearest classes
%             \item Generated n samples from Gaussian distribution based on the calibrated mean and co-variance
%             \item Repeat until the dataset is balanced
%         \end{enumerate}
%         \item Train the classifier based on this balanced dataset for 150 epochs (CIFAR100), 111 epochs (ImageNet)
%     \end{enumerate}
% \end{enumerate}

\section{Experiments}
\label{sec:experiments}

In this section, we start by presenting a brief overview of the dataset and implementation details for our methods.
We present some ablation studies comparing different aspects of the model, and finally perform a thorough comparison of our work against state-of-the-art approaches. 
We also include some analysis to obtain a better understanding of the feature generation process.

\subsection{Datasets and Experimental Setup}
\label{subsec:eval:data}
We benchmark our proposed method across two datasets, CIFAR-100-LT and mini-ImageNet-LT.

\paragraph{CIFAR-100-LT}
CIFAR-100 is a balanced dataset containing 60K images from 100 categories, with 50K in train and 10K in validation.
We use the synthetically created long-tail variants that have also been used by previous works~\cite{cao2019ldam,zhou2020bbn}.
There are three versions of the CIFAR-100-LT dataset, with the imbalance factor ($N_1 / N_K$) of $\{10, 50, 100\}$.
An exponential decay is used for determining $N_k$ in between.
Note that a higher imbalance factor mimics a stronger long-tail problem and is typically more challenging - we will see this through the performance of baseline models.

As the synthetically created variants involve randomly selecting images, all experiments are replicated with 3 seeds where different sets of images are selected.
The average performance over all seeds is reported when not mentioned otherwise.
As we will see in the ablation studies, this random sub-sampling often leads to a high variance, but we observe consistent improvements by using our approach.
For evaluation, we use the entire balanced validation set of 10K images, 100 samples for each of the 100 categories.

\paragraph{mini-ImageNet-LT}
mini-ImageNet was proposed by~\cite{vinyals2016matching} for few-shot learning evaluation, in an attempt to have a dataset like ImageNet while requiring fewer resources.
Similar to the statistics for CIFAR-100-LT with an imbalance factor of 100, we construct a long-tailed variant of mini-ImageNet that features all the 100 classes and an imbalanced training set with $N_1 = 500$ and $N_K = 5$ images.
% Then we synthetically create a variant with imbalance factor $N_1 / N_K = 100$. 
% Overall, it has 10.847K images from 100 categories, with maximally 500 images per class and minimally 5 images per class. 
For evaluation, both the validation and test sets are balanced and contain 10K images, 100 samples for each of the 100 categories.
\\

We report performance as accuracy.
Note that average per-sample accuracy is equivalent to average per-class accuracy due to the balanced evaluation sets in both datasets.

\subsection{Implementation details}
\label{subsec:eval:details}

\paragraph{Implementation details for CIFAR100-LT}
We follow a similar setup to~\cite{zhou2020bbn}:
the backbone is a ResNet-32~\cite{he2016resnet} trained by mini-batch stochastic gradient descent (SGD) with a momentum of 0.9 and a batch size of 128.
For a fair comparison, we tune the other parameters to match the baseline accuracy of the rest of the reported works.
The learning rate is decayed by a cosine scheduler~\cite{coslr2016sgdr} from 0.2 to 0.0 in 150 epochs, and a weight decay of $5 \times 10^{-5}$ is used.
% \todor{Add baseline matching ablations in ablations}.
We train our models on a single NVIDIA 1080Ti GPU.

\paragraph{Implementation details for mini-ImageNet-LT}
We follow a similar setup to~\cite{tang2020gbm, kang2019decoupling}: the backbone is a ResNeXt-50 trained by minibatch stochastic gradient descent (SGD) with a momentum of 0.9 for 111 epochs on a batch size of 512.
Through hyperparameter tuning, we found that weight decay of $5 \times 10^{-4}$ with a constant learning rate of 0.01 provides a better performing baseline.
We train all the models on a NVIDIA Tesla P100 GPU.
\\
% For the \ours{} feature generation, we use the following hyperparameters:
% Tukey $\lambda$ is chosen as 0.9 (1 for imbalance 100);
% $M$, the number of categorical numbers is 3 (2 for imbalance 10); and
% $\alpha$ is 0 (0.2 for imbalance 50).
% Subsequently, the classifier is trained with a constant learning rate of 0.001.

For the \ours{} feature generation, we use the hyperparameters as stated in Table~\ref{table:hyperparam}. 
Subsequently, the classifier is re-trained with a constant learning rate of 0.001 for CIFAR-100-LT and 0.01 for mini-ImageNet-LT.
While the weight decay for retraining the classifier remains the same as the backbone for CIFAR-100-LT, we found that decreasing the weight decay from $5 \times 10^{-4}$ to $5 \times 10^{-5}$ improves performance for mini-ImageNet-LT.

\begin{table}[t]
\caption{Hyperparameters used in \ours{} feature generation across various imbalance ratios and datasets.}
\label{table:hyperparam}
\begin{tabular}{rcccc}
\toprule
\multirow{3}{*}{Hyperparameters}& \multicolumn{3}{c}{CIFAR-100-LT} &\multirow{1}{*}{mini-}
\\
 & \multicolumn{3}{c}{Imbalance Ratios} &\multirow{1}{*}{-ImageNet-}\\
                            & 100   & 50    & 10   &\multirow{1}{*}{-LT}\\
\midrule
$\lambda$ (Tukey value)     & 1.0	& 0.9 & 0.9 & 1.0\\
$M$ (Number of categories)        & 3	& 2 & 2 & 2\\
$\alpha$ (Spread of generated features)         & 0.0	& 0.2 & 0.0 & 0.1\\

\bottomrule
\end{tabular}
\end{table}

% \todo{Rahul, add approx. time taken for generation, with details - num classes, num samples, etc.}

% \paragraph{Implementation details for ImageNet-LT}
% We follow a similar setup to \cite{tang2020gbm, kang2019decoupling}: the backbone is a ResNeXt-50 trained by minibatch stochastic gradient descent (SGD) with a momentum of 0.9 and weight decay of $2 \times 10^{-5}$ on a batch size of 512.
% The learning rate is decayed by a cosine scheduler from 0.2 to 0.0 in 111 epochs to match the baseline with the other reported works. \todor{Add baseline matching ablations in ablations}.
% We train all the models on a NVIDIA Tesla P100 GPU.

\subsection{Ablation studies on CIFAR-100-LT}
\label{subsec:eval:ablation}

We present a few ablation studies to compare various aspects of our proposed method.
Note that we do all the necessary ablations only with CIFAR-100-LT and use the findings to guide our decisions on mini-ImageNet-LT.

\paragraph{$\ell_2$-normalization}
\begin{table}[t]
\caption{Ablation study comparing the impact of $\ell_2$-normalization of features prior to \ours{}.
Backbone: CosineCE,
Feature generation: TailCalib.}
\label{table:cifar100-l2norm}
\begin{tabular}{rccc}
\toprule
\multirow{2}{*}{Method} & \multicolumn{3}{c}{Imbalance Ratios} \\
                            & 100   & 50    & 10   \\
\midrule
without $\ell_2$-normalization       & 42.62	& 49.17 & 58.12 \\
with $\ell_2$-normalization          & {\bf 43.03}	& 49.18 & {\bf 58.60} \\
\bottomrule
\end{tabular}

\end{table}

As indicated in the last paragraph of Sec.~\ref{subsec:tailcalib}, when using CosineCE, we observed that it is better to $\ell_2$-normalize the feature vectors $\zvec_i$ before and after the generation.
We show the impact of this normalization in Table~\ref{table:cifar100-l2norm}.
Imbalance factors 100 and 10 see a small but consistent improvement across all three seeds, pointing to the relevance of this modification:
+0.21\%, +0.64\%, and +0.38\% for imbalance 100 and 
+0.77\%, +0.4\%, and +0.29\% for imbalance 10.
For imbalance 50, we see a small improvement for 2 of 3 seeds:
+0.12\%, +0.09\%, and -0.19\%, resulting in an overall negligible performance difference.

\begin{figure*}[t]
\centering
\includegraphics[width=0.33\linewidth]{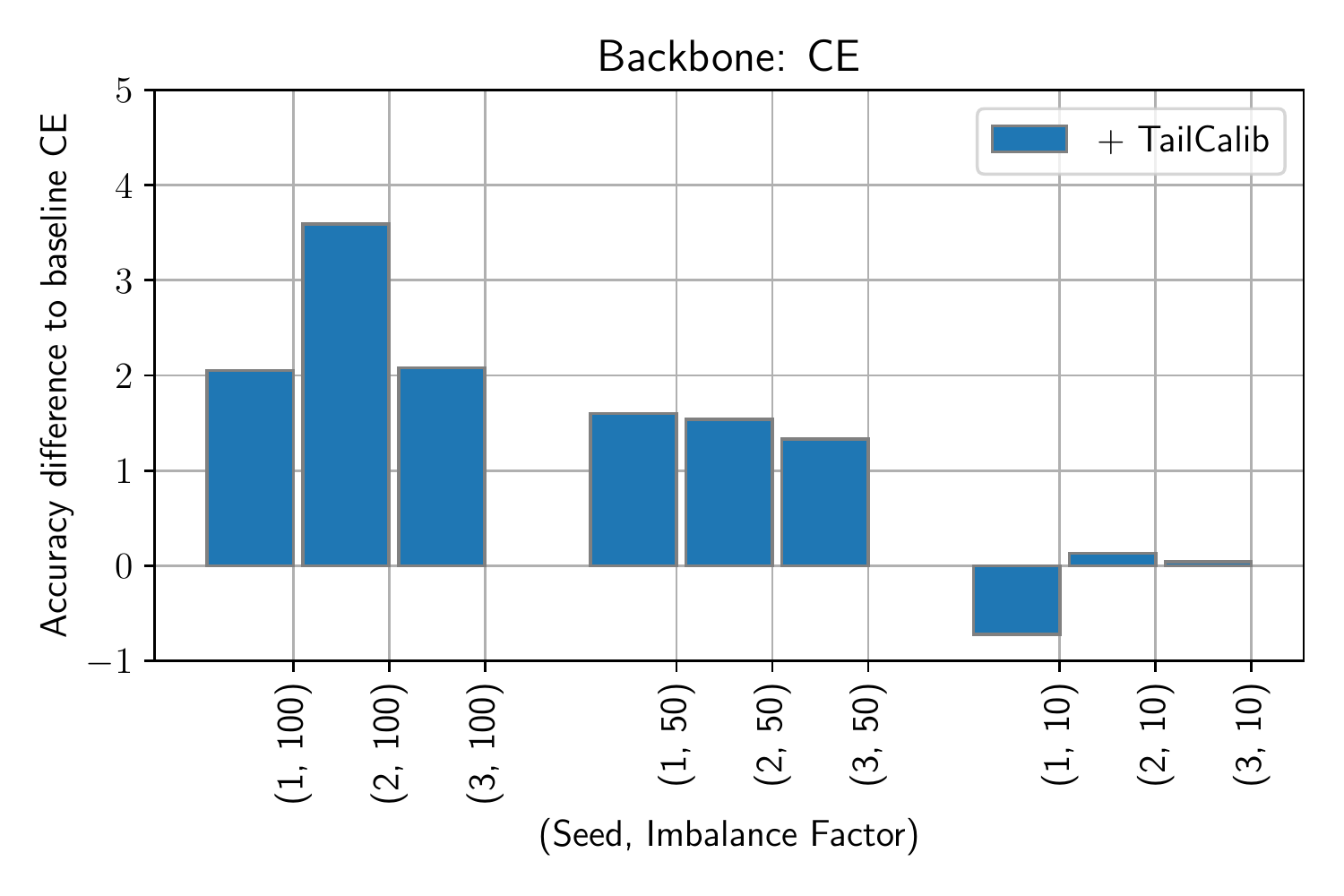}
\includegraphics[width=0.33\linewidth]{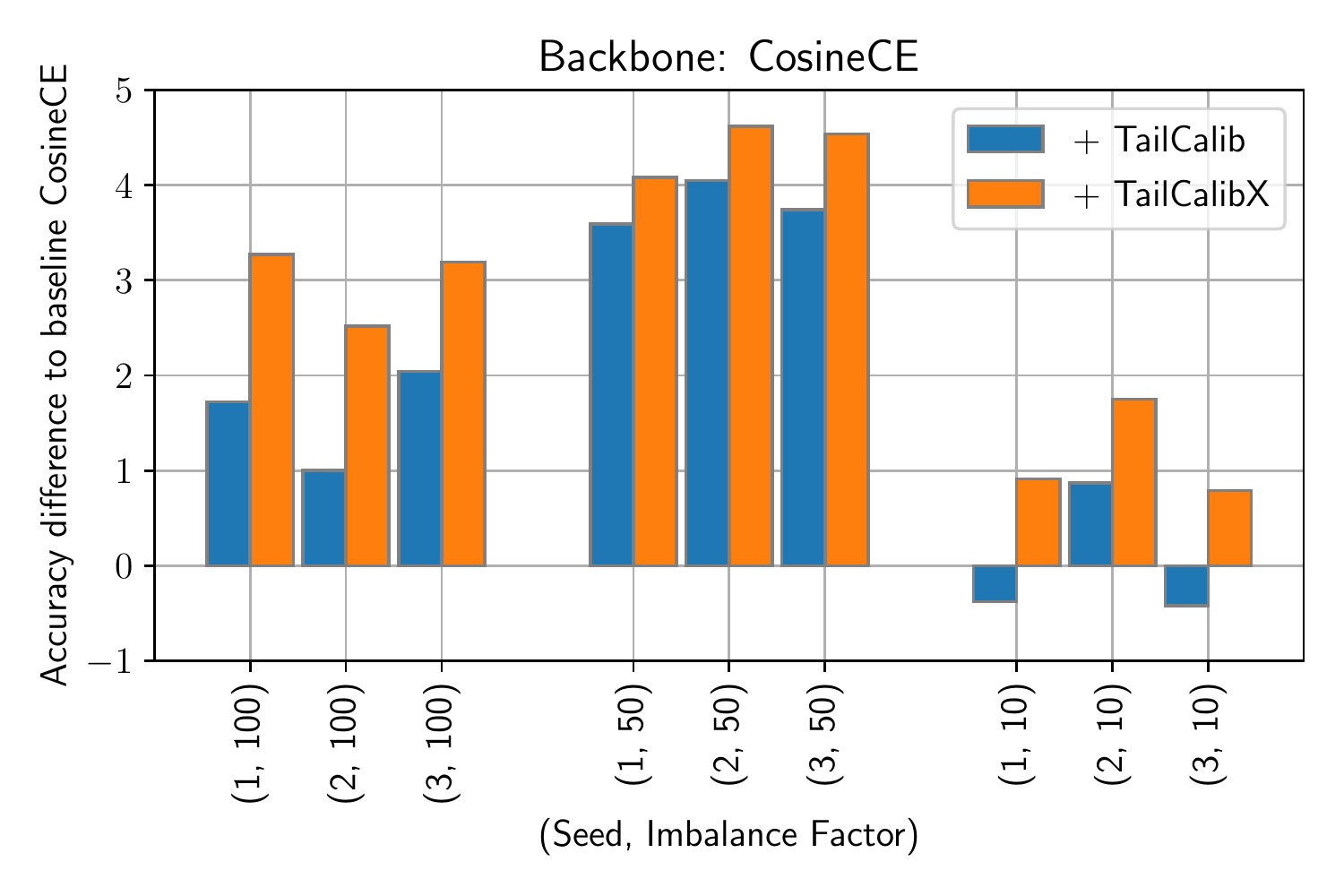}
\includegraphics[width=0.33\linewidth]{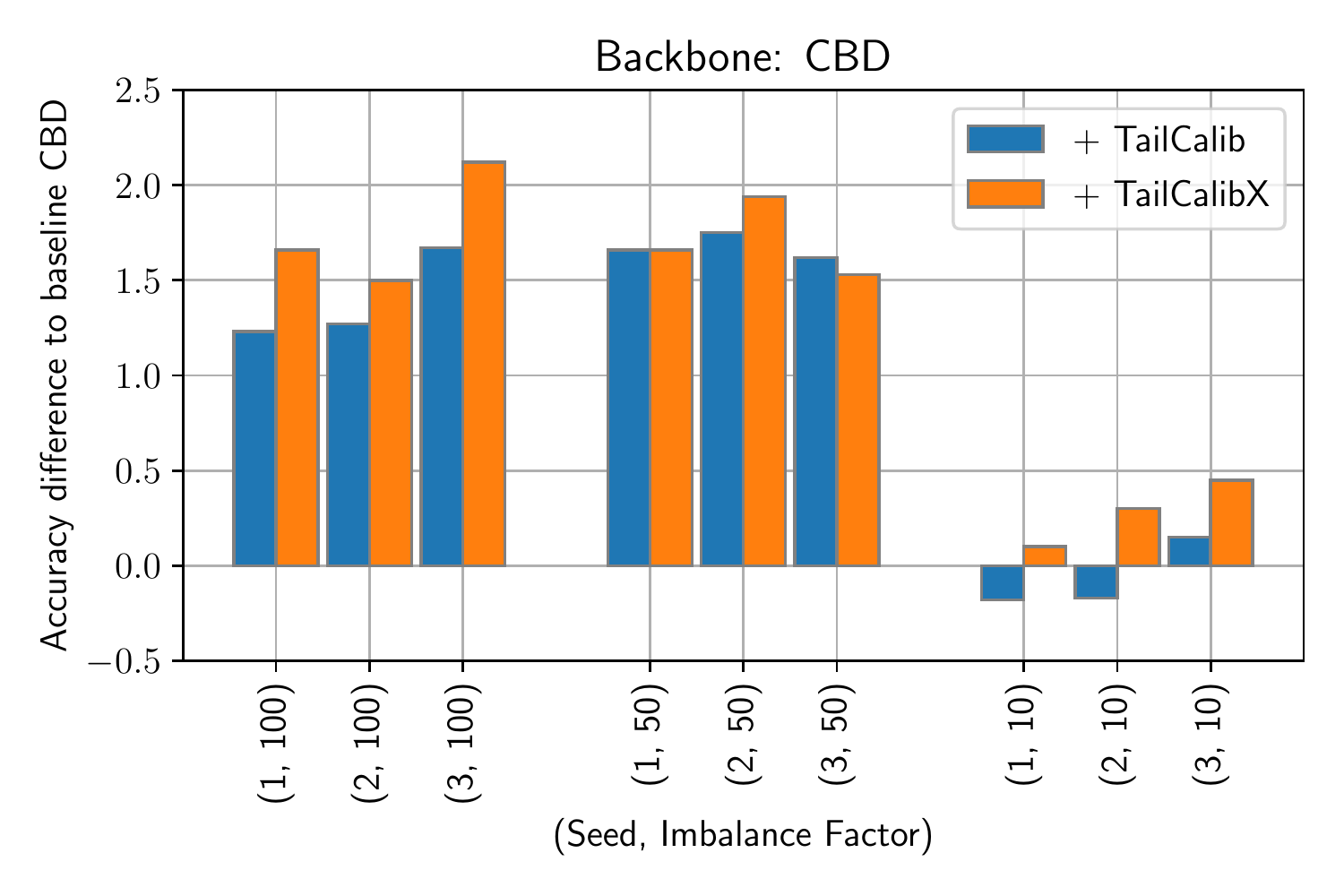}
\vspace{-5mm}
\caption{Performance improvement in absolute percentage points across TailCalib and TailCalibX for all three backbones: CE (left), CosineCE (middle), CBD (right).
On the x-axis, we indicate the seed and imbalance factor of each run.
Please refer to Table~\ref{table:cifar100-backbones} for averaged performance scores across the 3 seeds.}
\label{fig:cifar100-backbones-seeds}
\end{figure*}

\paragraph{Backbones}
\begin{table}[t]
\caption{Ablation study comparing the impact of backbones: CE vs. CosineCE vs. CBD.
We also see that generating features multiple times can be beneficial for training: TailCalib vs. TailCalibX.}
\label{table:cifar100-backbones}
\begin{tabular}{rlccc}
\toprule
\multirow{2}{*}{Backbone} & Feature & \multicolumn{3}{c}{Imbalance Ratios} \\
& Generation            & 100   & 50    & 10   \\
\midrule
\multirow{3}{*}{CE}
    & -                 & 39.89 & 44.88 & {\bf 57.00} \\
    & TailCalib         & {\bf 42.28} & {\bf 46.37} & 56.81 \\
    % & TailCalibX         & {\bf 42.86} & FILL & FILL \\
\midrule
\multirow{3}{*}{CosineCE}
    & -                 & 41.44 & 45.72 & 58.58 \\
    & TailCalib         & 43.03 & 49.18 & 58.60 \\
    & TailCalibX        & {\bf 44.44} & {\bf 49.80} & {\bf 59.73} \\
\midrule
\multirow{3}{*}{CBD~\cite{iscen2021cbd}}
    & -                 & 44.83 & 49.19 & 60.85 \\
    & TailCalib         & 46.22 & {\bf 50.87} & 60.78 \\
    & TailCalibX        & {\bf 46.59} & {\bf 50.90} & {\bf 61.13} \\
\bottomrule
\end{tabular}
\end{table}

We presented three main strategies to train our backbone in Sec.~\ref{subsec:backbone}.
Table~\ref{table:cifar100-backbones} reports performance when applying \ours{} (abbreviated as TailCalib) for features precomputed from various backbones (averaged across 3 seeds).
Firstly, note that CosineCE outperforms CE, while CBD with teacher-student distillation outperforms both.
\ours{} shows a consistent improvement over backbones, especially for the higher imbalance factors of 50 and 100.

\paragraph{Training with multiple rounds of generation}
As the feature generation process is quite fast, we can afford to generate samples on-the-fly for each epoch during the training of the classifier.
In particular, we propose \emph{TailCalibX}, denoting that TailCalib is employed multiple times.
Specifically, we generate a set of features once every epoch and use them to train the classifier.
We show the results for this experiment in Table~\ref{table:cifar100-backbones}.
TailCalibX provides incremental performance boosts over TailCalib.

\paragraph{Trends across different seeds}
The variation across seeds is typically high as changing the seed not only changes the random initialization of the base network (this has a small impact), but also leads to the selection of a different subset of CIFAR-100 images for creating the synthetically imbalanced training data (this has a large impact).
This is especially true for imbalance factor 50 and 10; we suspect that as there are very few training samples in the tail categories for imbalance factor 100, which samples are selected does not matter as much.
For example, the accuracy of the CE baseline across 3 seeds is (43.8\%, 44.8\%, 46.1\%) for imbalance factor 50.

Fig.~\ref{fig:cifar100-backbones-seeds} shows the absolute percentage points performance improvement over the specific backbone for each seed.
We observe consistent performance improvement over progressively harder backbones by applying TailCalib or TailCalibX across various seeds.

We can also see that while TailCalib may fail to provide performance improvements for low imbalance factors (CosineCE and CBD, imbalance 10), TailCalibX always provides a small boost.
This indicates that the feature generation process may not be very reliable for datasets with a small long-tail effect, but the classifier can still extract meaningful information from multiple rounds of generation.

\subsection{Comparison to state-of-the-art}
\label{subsec:eval:sota}

We compare our proposed approach against several previous works grouping them in a manner similar to our related work in Sec.~\ref{sec:relatedwork}.
% based on loss or sample/class re-weighting mechanisms, decoupling of classifier and representation learning, distillation methods, and sample or feature generation.

\begin{table}[t]
\tabcolsep=0.14cm
\caption{Comparison of our approach (TailCalib, TailCalibX) against previous works on CIFAR-100-LT.
Methods with a * indicate results produced by our re-implementations, please see the text for details about the implementation.
Best results are highlighted in bold, next best and notable results with underline.}
\label{table:cifar100-sota}
\begin{tabular}{rlccc}
\toprule
\multirow{2}{*}{Type} & \multirow{2}{*}{Method} & \multicolumn{3}{c}{Imbalance Ratios} \\
 &      & 100   & 50    & 10   \\
\midrule
\multirow{2}{*}{Baseline}
    & *CE
        & 39.89 & 44.88 & 57.00 \\
    & *CosineCE
        & 41.44 & 45.72 & 58.58 \\
\midrule
    & Focal Loss~\cite{lin2017focal}
        & 38.41 & 44.32 & 55.78 \\
    & Class-Balanced Focal~\cite{cui2019classbalanced}
        & 39.60 & 45.32 & 57.99 \\
    & L2RW~\cite{ren2018metareweight}
        & 38.90 & 46.83 & 52.12 \\
Loss or    & Meta-Weight Net~\cite{shu2019metaweightnet}
        & 41.61 & 45.66 & 58.91 \\
Re-weighting    & Domain Adaptation~\cite{jamal2020domainadaptation}
        & 39.31 & 48.53 & 59.58 \\
    & LDAM-DRW~\cite{cao2019ldam}
        & 42.04 & -     & 58.71 \\
    & Logit adjustment~\cite{menon2021logitadjustment}
        & 43.89 & -     & -     \\
    & GBM~\cite{tang2020gbm}
        & 44.1  & 50.3  & 59.6  \\
\midrule    
\multirow{2}{*}{Decouple}
    & *cRT~\cite{kang2019decoupling}
        & 42.63 & 47.26 & 57.61 \\
    & BBN~\cite{zhou2020bbn}
        & 42.56 & 47.02 & 59.12 \\
\midrule
\multirow{2}{*}{Distillation}
    & LFME~\cite{xiang2020lfme}
        & 42.3  & -     & -     \\
    & *CBD~\cite{iscen2021cbd}
        & {\ul 44.83} & {\ul 49.19} & {\ul 60.85} \\
\midrule
\multirow{4}{*}{Generation}
    & M2M~\cite{kim2020m2m}
        & 42.9  & -     & 58.2 \\
    & mixup~\cite{zhang2017mixup}
        & 39.54 & 44.99 & 58.02 \\
    & Manifold mixup~\cite{verma2019manifoldmixup}
        & 38.25 & 43.09 & 56.55 \\
    & *MODALS~\cite{cheung2021modals}
        & 42.813 & 47.57 & 58.53 \\
    % & *MODALS (lr 0.2)~\cite{cheung2021modals}
    %     & 42.29 & 47.01 & 58.25 \\
\midrule
\multirow{3}{*}{*Ours}
    & CosineCE + TailCalib
        & 43.03 & 49.18 & 58.60 \\
    & CosineCE + TailCalibX
        & {\ul 44.44} & {\ul 49.80} & {\ul 59.73} \\
    & CBD + TailCalibX
        & \textbf{46.59} & \textbf{50.90} & \textbf{61.13} \\
\bottomrule
\end{tabular}
\end{table}

\begin{table}[t]
\tabcolsep=0.11cm
\caption{Comparison of our approach (TailCalib, TailCalibX) against previous works on mini-ImageNet-LT.
Since we introduce using mini-ImageNet in a long-tail variant, all methods are our re-implementations.
Best results are highlighted in bold, next best and notable results with underline.}
\label{table:miniimagenet-sota}
\begin{tabular}{rlcccc}
\toprule
\multirow{1}{*}{Type} & \multirow{1}{*}{Method} 
 &      \multirow{1}{*}{Many}   & \multirow{1}{*}{Mid}    & \multirow{1}{*}{Few}     & \multirow{1}{*}{All}   \\
\midrule
\multirow{2}{*}{Baseline}
    & CE
        & 57.77 & 29.42 & 8.06 & 32.94 \\
    & CosineCE
        & 65.45 & 29.37 & 8.6 & 35.77 \\
\midrule    
\multirow{1}{*}{Decouple}
    & cRT~\cite{kang2019decoupling}
        & 66.00 & 31.99 & {\ul 29.3} & {\ul 43.09} \\
\midrule
\multirow{1}{*}{Distillation}
    & CBD~\cite{iscen2021cbd}
        & 67.34 &  \textbf{35.00} & 23.06 & 42.74  \\
\midrule
\multirow{2}{*}{Generation}
    & MODALS~\cite{cheung2021modals}
        & 66.62 & 28.65 & 11.06 & 36.67 \\
    & mixup~\cite{zhang2017mixup}
        & 64.71 & 27.51 & 8.1 & 34.72 \\
\midrule
\multirow{2}{*}{Ours}
    & CosineCE + TailCalib
        & {\ul 68.05} & 32.59 & 25.13 & 42.77 \\
    % & CosineCE + TailCalib (5e-3)
    %     & 64.4 & 29.2 & 25.23 & 40.33 \\
    & CosineCE + TailCalibX
        & \textbf{68.3} & {\ul 34.3} & \textbf{29.4} & \textbf{44.73} \\
\bottomrule
\end{tabular}

\end{table}

Results for most methods are taken from the respective papers or BBN~\cite{zhou2020bbn}.
For the methods that did not report results on CIFAR-100-LT or mini-ImageNet-LT, we present additional details of our re-implementation.

\begin{itemize}[leftmargin=0.4cm]
\item For CE and CosineCE, we choose hyperparameters (learning rate scheduler, number of epochs, etc.) to match the performance of CE on a previously reported work, BBN~\cite{zhou2020bbn}.
\item cRT~\cite{kang2019decoupling} does not provide results on the CIFAR-100-LT and mini-ImageNet-LT datasets. We use their publicly available code to generate results for both the datasets.
\item CBD is a recent, and to the best of our knowledge, unpublished work~\cite{iscen2021cbd}.
Through email conversations with the authors, we verified our implementation and present results with a single teacher trained on the both datasets.
\item MODALS~\cite{cheung2021modals} was trained with the Gaussian noise augmentation.
We choose $\lambda = 0.01$ after performing a sweep across $[0.1, 0.01, 0.001]$.
\item For mixup~\cite{zhang2017mixup}, we choose $\alpha = 0.01$ after performing a sweep across $[0.01,0.1,0.2, 0.3, 0.5, 0.7]$.
\end{itemize}

\begin{figure*}[t]
\centering
\includegraphics[height=0.34\linewidth]{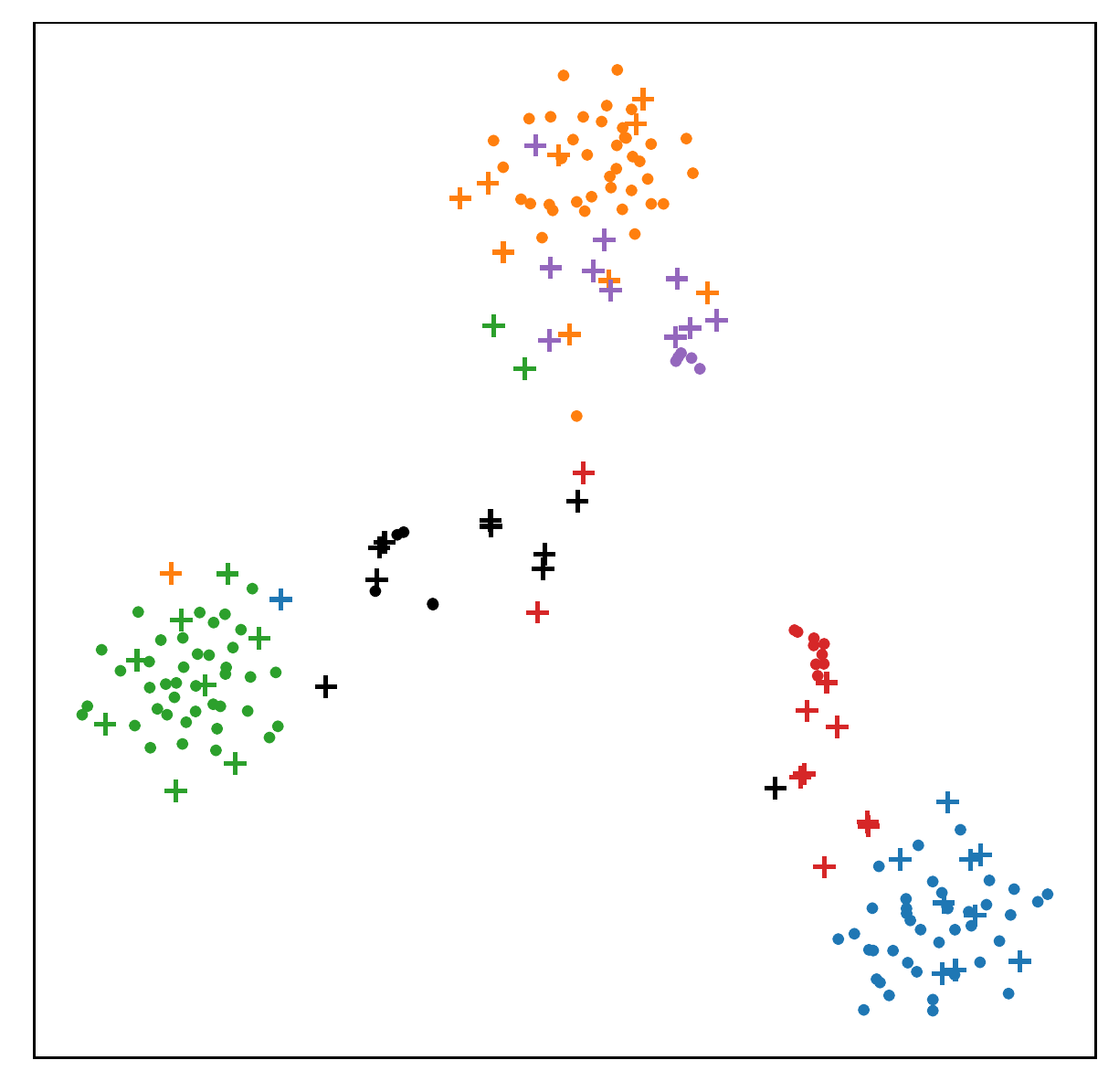}
\includegraphics[height=0.34\linewidth]{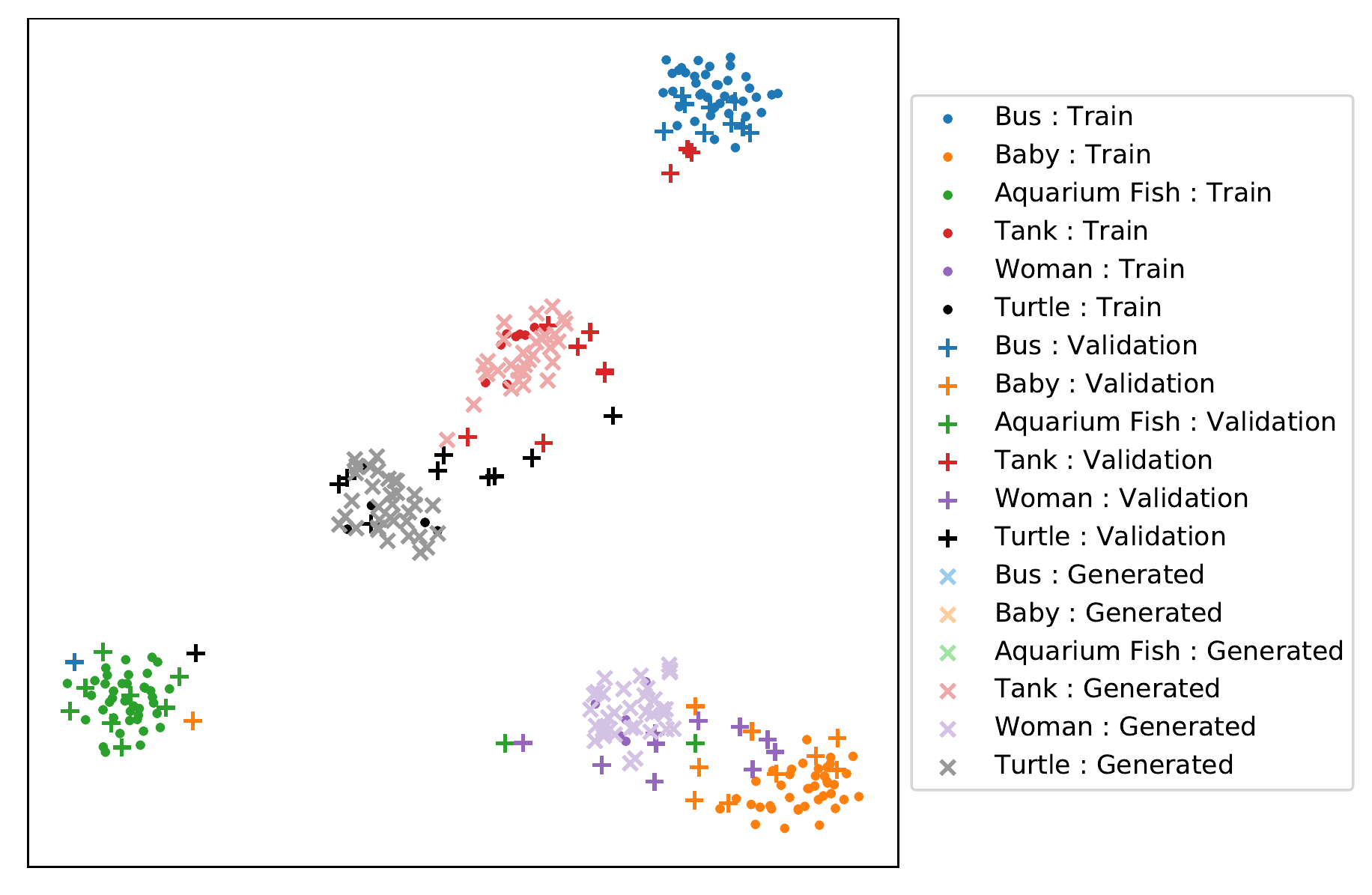}
\vspace{-1mm}
\caption{t-SNE visualization of a few head and tail classes from CIFAR-100-LT.
The plot on the {\ul left is before generation}, and the plot on the {\ul right is after generation}.
We show 10 validation samples for each class and limit to 40 training + generated samples for ease of interpretation.
{\ul Markers:} $\bf \cdot$ (dot) indicate training samples; $+$ (plus) are validation samples; and $\times$ (cross) are generated features also shown with a lighter version of the base color (\eg~black-gray for \emph{turtle}).
{\ul Head categories:} \emph{bus} (273), \emph{baby} (455), \emph{aquarium fish} (477).
{\ul Tail categories:} \emph{tank} (9), \emph{woman} (5), \emph{turtle} (6).
Class size is indicated in parenthesis.
Please refer to the text, Sec.~\ref{subsec:eval:analysis} for a more thorough discussion.
Best seen in colour.}
\label{fig:tsne1}
\end{figure*}

% \begin{figure*}[t]
% \centering
% \includegraphics[width=0.45\linewidth]{figures/tsne2_before_nolegend.pdf}
% \includegraphics[width=0.545\linewidth]{figures/tsne2_after.pdf}
% \caption{A second example of t-SNE visualization of a few head and tail classes from CIFAR-100-LT, showing 10 validation samples for each class and limited to 40 training + generated samples for ease of interpretation.
% {\ul Markers:} $\bf \cdot$ (dot) indicate training samples; $+$ (plus) are validation samples; and $\times$ (cross) are generated features.
% {\ul Head categories:} \emph{bear} (434), \emph{beaver} (415), \emph{bed} (396).
% {\ul Tail categories:} \emph{otter} (55), \emph{tiger} (88), \emph{table} (84).
% Class size is indicated in parenthesis.
% The plot on the {\ul left is before generation}, and the plot on the {\ul right is after generation}.
% Please refer to the text, Sec.~\ref{subsec:eval:analysis} for a more thorough discussion.
% Best seen in colour.}
% \label{fig:tsne2}
% \end{figure*}

\paragraph{CIFAR-100-LT (Table~\ref{table:cifar100-sota})}
Firstly, note that the CosineCE training paradigm is superior to CE, and in fact, outperforms several loss re-weighting based methods.
CosineCE + TailCalibX outperforms all previous works ignoring distillation.
In fact, it achieves performance close to CBD~\cite{iscen2021cbd}  (especially on imbalance factors 100 and 50) that trains two networks - a teacher and student, and benefits from the advantages of ensembling.
Finally, applying TailCalibX on top of a trained CBD model results in further performance improvements of 1-2\% notably for imbalance factors 100 and 50.

\paragraph{mini-ImageNet-LT (Table~\ref{table:miniimagenet-sota})}
TailCalibX outperforms all previous works.
As compared to the CosineCE baseline, not only does it improve the accuracy for tail classes (\textit{few}) by over 20\%, we also see performance improvements for both the middle classes (\textit{mid}, +4.93\%) and head classes (\textit{many}, +2.85\%) as well.
Surprisingly, CBD performs worse than classifier re-training (cRT), hence, we omit evaluation of \ours{} as applied to a backbone trained with CBD.

\subsection{Analysis}
\label{subsec:eval:analysis}

\paragraph{tSNE visualization}
Fig.~\ref{fig:tsne1} shows a few feature embeddings computed using t-SNE~\cite{maaten2008tsne} for head and tail categories from the CIFAR-100-LT dataset.
For visual clarity, we show 10 validation samples for each class and up to 40 training + generated samples.
We randomly choose 3 pairs of visually similar categories: \emph{tank} and \emph{bus}, \emph{woman} and \emph{baby}, and \emph{aquarium fish} and \emph{turtle}, such that one of them is from the bottom 15 tail categories, while the second belongs to the top 15 head categories.

In the left plot, before feature generation, we see that the validation samples ($+$) from the tail categories are often confused with visually similar and larger head classes:
(i) five samples of the \emph{tank} (red) are located close to the \emph{bus} (blue) cluster; and
(ii) roughly 7 of 10 samples of the \emph{woman} category (purple) are potentially confused as belonging to the \emph{baby} class (orange).
In the right plot, we generate features using \ours{} and re-compute the t-SNE embeddings.
We observe a noticeable reduction in errors, even on this randomly picked toy sample.
Now, 3 of 5 \emph{tank} samples are close to the \emph{bus} cluster, and 4 of 7 \emph{woman} samples are close to the \emph{baby} cluster, while the others are more easily separable.

\paragraph{Class nearest neighbors}
\begin{table}[t]
\caption{Two most frequently chosen nearest neighbor categories to calibrate the distributions for generating features for 20 tail categories of CIFAR-100-LT dataset.}
\label{table:cifar100-nn}
\begin{tabular}{lcl}
\toprule
Tail category       & \# Train samples & Nearest neighbors \\
\midrule
% {\bf squirrel}      & 12 & dinosaur, elephant \\
% {\bf streetcar}     & 11 & train, bus \\
% {\bf sunflower}     & 11 & caterpillar, poppy \\
% {\bf sweet\_pepper} & 10 & rose, poppy \\
% {\bf table}         & 10 & couch, lobster \\
{\bf tank}          &  9 & tractor, pickup\_truck \\
{\bf telephone}     &  9 & streetcar, keyboard \\
{\bf television}    &  8 & couch, lamp \\
{\bf tiger}         &  8 & leopard, lion \\
{\bf tractor}       &  7 & lawn\_mower, train \\
{\bf train}         &  7 & tractor, road \\
{\bf trout}         &  7 & dinosaur, caterpillar \\
{\bf tulip}         &  6 & rabbit, cattle \\
{\bf turtle}        &  6 & telephone, ray \\
{\bf wardrobe}      &  6 & television, skyscraper \\
{\bf whale}         &  6 & otter, seal \\
{\bf willow\_tree}  &  5 & rabbit, oak\_tree \\
{\bf wolf}          &  5 & rabbit, possum \\
{\bf woman}         &  5 & man, girl \\
{\bf worm}          &  5 & ray, woman \\
\bottomrule
\end{tabular}
\end{table}

Recall, to generate the calibrated tail distributions, we pick $M$ categories as nearest neighbors for each feature $\zvec_{ik}$.
We analyze the classes that are most commonly used as nearest neighbors for the bottom 15 tail classes of the CIFAR-100-LT dataset in Table~\ref{table:cifar100-nn}.
The table shows the nearest neighbors for imbalance factor 100 and $M = 3$.
We observe that the class for which we wish to generate samples is often the nearest centroid and is ignored in the table for brevity.

This split of CIFAR-100-LT is sorted alphabetically, \ie~the classes with letter \emph{a} belong to the head, and classes with letters towards the end of the alphabet form the tail.
We see that nearest neighbors need not belong to the head classes, even though it may be common.
For example, we see that \emph{tiger} distribution draws from the \emph{lion} class, or \emph{television} from \emph{couch}, possibly due to the fact that images with a television have additional furniture in them.

\section{Conclusion and Future work}
\label{sec:conclusion}

Learning to cope with long-tail distributions is an important problem for machine learning.
Inspired by a recent work on few-shot learning~\cite{yang2021freelunch}, we explored one facet of this challenging task: feature generation as a means to balance the training data seen by a classifier.
We analyzed the efficacy of distribution calibration and empirically validated that it can lead to substantial performance improvements.
In this process, we achieved a new state-of-the-art on two synthetic datasets: CIFAR-100-LT and mini-ImageNet-LT, especially demonstrating that \ours{} is orthogonal to, and can be combined with, other works such as CBD~\cite{iscen2021cbd}, to yield additional improvements.

\paragraph{Future work}
In the future, we wish to analyze the feature generation process in a more thorough manner.
In particular, we hope to discover the underlying class-specific manifolds on which the features would be expected to lie.
While our initial attempts at evaluating \ours{} on ImageNet did not yield significant improvement, we would like to extend our evaluation to other long-tail datasets, both synthetic: ImageNet-LT~\cite{liu2019oltr} and Places-LT~\cite{liu2019oltr}, and natural: iNaturalist~\cite{vanhorn2018inaturalist} or FineFoods dataset~\cite{finefoodsworkshop}. 
Another possibility is to exploit the hierarchy of categories and learn appropriate models such as hyperbolic representations \cite{pmlr-v97-law19a} or their generalization to pseudo-Riemannian manifolds of constant nonzero curvature \cite{NEURIPS2020_123b7f02}.

\paragraph{Acknowledgments}
This work has been partly supported by the funding received from DST through the IMPRINT program (IMP / 2019 / 000250).
\bibliographystyle{ACM-Reference-Format}
\bibliography{longstrings,refs}


\begin{thebibliography}{57}

%%% the Web address.

\ifx \showCODEN    \undefined \def \showCODEN     #1{\unskip}     \fi
\ifx \showDOI      \undefined \def \showDOI       #1{#1}\fi
\ifx \showISBNx    \undefined \def \showISBNx     #1{\unskip}     \fi
\ifx \showISBNxiii \undefined \def \showISBNxiii  #1{\unskip}     \fi
\ifx \showISSN     \undefined \def \showISSN      #1{\unskip}     \fi
\ifx \showLCCN     \undefined \def \showLCCN      #1{\unskip}     \fi
\ifx \shownote     \undefined \def \shownote      #1{#1}          \fi
\ifx \showarticletitle \undefined \def \showarticletitle #1{#1}   \fi
\ifx \showURL      \undefined \def \showURL       {\relax}        \fi
% invisible to TeX
\providecommand\bibfield[2]{#2}
\providecommand\bibinfo[2]{#2}
\providecommand\natexlab[1]{#1}
\providecommand\showeprint[2][]{arXiv:#2}

\bibitem[\protect\citeauthoryear{Acuna, Zhang, Law, and Fidler}{Acuna
  et~al\mbox{.}}{2021}]%
        {acuna2021fdomainadversarial}
\bibfield{author}{\bibinfo{person}{David Acuna}, \bibinfo{person}{Guojun
  Zhang}, \bibinfo{person}{Marc~T. Law}, {and} \bibinfo{person}{Sanja Fidler}.}
  \bibinfo{year}{2021}\natexlab{}.
\newblock \showarticletitle{f-Domain Adversarial Learning: Theory and
  Algorithms}. In \bibinfo{booktitle}{\emph{International Conference on Machine
  Learning (ICML)}}.
\newblock


\bibitem[\protect\citeauthoryear{Banerjee, Dhillon, Ghosh, Sra, and
  Ridgeway}{Banerjee et~al\mbox{.}}{2005}]%
        {banerjee2005clustering}
\bibfield{author}{\bibinfo{person}{Arindam Banerjee},
  \bibinfo{person}{Inderjit~S Dhillon}, \bibinfo{person}{Joydeep Ghosh},
  \bibinfo{person}{Suvrit Sra}, {and} \bibinfo{person}{Greg Ridgeway}.}
  \bibinfo{year}{2005}\natexlab{}.
\newblock \showarticletitle{Clustering on the Unit Hypersphere using von
  Mises-Fisher Distributions}.
\newblock \bibinfo{journal}{\emph{Journal of Machine Learning Research}}
  \bibinfo{volume}{6}, \bibinfo{number}{9} (\bibinfo{year}{2005}).
\newblock


\bibitem[\protect\citeauthoryear{Cao, Wei, Gaidon, Arechiga, and Ma}{Cao
  et~al\mbox{.}}{2019}]%
        {cao2019ldam}
\bibfield{author}{\bibinfo{person}{Kaidi Cao}, \bibinfo{person}{Colin Wei},
  \bibinfo{person}{Adrien Gaidon}, \bibinfo{person}{Nikos Arechiga}, {and}
  \bibinfo{person}{Tengyu Ma}.} \bibinfo{year}{2019}\natexlab{}.
\newblock \showarticletitle{{Learning Imbalanced Datasets with
  Label-Distribution-Aware Margin Loss}}. In \bibinfo{booktitle}{\emph{Advances
  in Neural Information Processing Systems (NeurIPS)}}.
\newblock


\bibitem[\protect\citeauthoryear{Cao, Law, and Fidler}{Cao
  et~al\mbox{.}}{2020}]%
        {Cao2020A}
\bibfield{author}{\bibinfo{person}{Tianshi Cao}, \bibinfo{person}{Marc~T Law},
  {and} \bibinfo{person}{Sanja Fidler}.} \bibinfo{year}{2020}\natexlab{}.
\newblock \showarticletitle{A Theoretical Analysis of the Number of Shots in
  Few-Shot Learning}. In \bibinfo{booktitle}{\emph{International Conference on
  Learning Representations (ICLR)}}.
\newblock


\bibitem[\protect\citeauthoryear{Chawla, Bowyer, Hall, and Kegelmeyer}{Chawla
  et~al\mbox{.}}{2002}]%
        {chawla2002smote}
\bibfield{author}{\bibinfo{person}{Nitesh~V. Chawla}, \bibinfo{person}{Kevin~W.
  Bowyer}, \bibinfo{person}{Lawrence~O. Hall}, {and} \bibinfo{person}{W.~Philip
  Kegelmeyer}.} \bibinfo{year}{2002}\natexlab{}.
\newblock \showarticletitle{{SMOTE: Synthetic Minority Over-sampling
  Technique}}.
\newblock \bibinfo{journal}{\emph{Journal of Artificial Intelligence Research
  (JAIR)}}  \bibinfo{volume}{16} (\bibinfo{year}{2002}),
  \bibinfo{pages}{321--357}.
\newblock


\bibitem[\protect\citeauthoryear{Cheung and Yeung}{Cheung and Yeung}{2021}]%
        {cheung2021modals}
\bibfield{author}{\bibinfo{person}{Tsz-Him Cheung} {and}
  \bibinfo{person}{Dit-Yan Yeung}.} \bibinfo{year}{2021}\natexlab{}.
\newblock \showarticletitle{{MODALS: Modality-agnostic Automated Data
  Augmentation in the Latent Space}}. In
  \bibinfo{booktitle}{\emph{International Conference on Learning
  Representations (ICLR)}}.
\newblock


\bibitem[\protect\citeauthoryear{Chu, Bian, Liu, and Ling}{Chu
  et~al\mbox{.}}{2020}]%
        {chu2020feature}
\bibfield{author}{\bibinfo{person}{Peng Chu}, \bibinfo{person}{Xiao Bian},
  \bibinfo{person}{Shaopeng Liu}, {and} \bibinfo{person}{Haibin Ling}.}
  \bibinfo{year}{2020}\natexlab{}.
\newblock \showarticletitle{Feature space augmentation for long-tailed data}.
  In \bibinfo{booktitle}{\emph{European Conference on Computer Vision (ECCV)}}.
\newblock


\bibitem[\protect\citeauthoryear{Cui, Jia, Lin, Song, and Belongie}{Cui
  et~al\mbox{.}}{2019}]%
        {cui2019classbalanced}
\bibfield{author}{\bibinfo{person}{Yin Cui}, \bibinfo{person}{Menglin Jia},
  \bibinfo{person}{Tsung-Yi Lin}, \bibinfo{person}{Yang Song}, {and}
  \bibinfo{person}{Serge Belongie}.} \bibinfo{year}{2019}\natexlab{}.
\newblock \showarticletitle{Class-balanced loss based on effective number of
  samples}. In \bibinfo{booktitle}{\emph{Conference on Computer Vision and
  Pattern Recognition (CVPR)}}.
\newblock


\bibitem[\protect\citeauthoryear{Damien~Dablain}{Damien~Dablain}{2021}]%
        {dablain2021deepsmote}
\bibfield{author}{\bibinfo{person}{Nitesh V.~Chawla Damien~Dablain,
  Bartosz~Krawczyk}.} \bibinfo{year}{2021}\natexlab{}.
\newblock \showarticletitle{{DeepSMOTE: Fusing Deep Learning and SMOTE for
  Imbalanced Data}}. In \bibinfo{booktitle}{\emph{arXiv:2105.02340}}.
\newblock


\bibitem[\protect\citeauthoryear{Dhillon, Chaudhari, Ravichandran, and
  Soatto}{Dhillon et~al\mbox{.}}{2020}]%
        {dhillon2020fslbaseline}
\bibfield{author}{\bibinfo{person}{Guneet~Singh Dhillon},
  \bibinfo{person}{Pratik Chaudhari}, \bibinfo{person}{Avinash Ravichandran},
  {and} \bibinfo{person}{Stefano Soatto}.} \bibinfo{year}{2020}\natexlab{}.
\newblock \showarticletitle{{A Baseline for Few-Shot Image Classification}}. In
  \bibinfo{booktitle}{\emph{International Conference on Learning
  Representations (ICLR)}}.
\newblock


\bibitem[\protect\citeauthoryear{Gemmeke, Ellis, Freedman, Jansen, Lawrence,
  Moore, Plakal, and Ritter}{Gemmeke et~al\mbox{.}}{2017}]%
        {gemmeke2017audioset}
\bibfield{author}{\bibinfo{person}{Jort~F. Gemmeke}, \bibinfo{person}{Daniel
  P.~W. Ellis}, \bibinfo{person}{Dylan Freedman}, \bibinfo{person}{Aren
  Jansen}, \bibinfo{person}{Wade Lawrence}, \bibinfo{person}{R.~Channing
  Moore}, \bibinfo{person}{Manoj Plakal}, {and} \bibinfo{person}{Marvin
  Ritter}.} \bibinfo{year}{2017}\natexlab{}.
\newblock \showarticletitle{{Audio Set: An ontology and human-labeled dataset
  for audio events}}. In \bibinfo{booktitle}{\emph{International Conference on
  Audio, Speech, and Signal Processing (ICASSP)}}.
\newblock


\bibitem[\protect\citeauthoryear{Gidaris and Komodakis}{Gidaris and
  Komodakis}{2018}]%
        {gidaris2018cosinece}
\bibfield{author}{\bibinfo{person}{Spyros Gidaris} {and} \bibinfo{person}{Nikos
  Komodakis}.} \bibinfo{year}{2018}\natexlab{}.
\newblock \showarticletitle{{Dynamic Few-shot Visual Learning without
  Forgetting}}. In \bibinfo{booktitle}{\emph{Conference on Computer Vision and
  Pattern Recognition (CVPR)}}.
\newblock


\bibitem[\protect\citeauthoryear{Gu, Sun, Ross, Vondrick, Pantofaru, Li,
  Vijayanarasimhan, Toderici, Ricco, Sukthankar, Schmid, and Malik}{Gu
  et~al\mbox{.}}{2018}]%
        {gu2018ava}
\bibfield{author}{\bibinfo{person}{Chunhui Gu}, \bibinfo{person}{Chen Sun},
  \bibinfo{person}{David~A. Ross}, \bibinfo{person}{Carl Vondrick},
  \bibinfo{person}{Caroline Pantofaru}, \bibinfo{person}{Yeqing Li},
  \bibinfo{person}{Sudheendra Vijayanarasimhan}, \bibinfo{person}{George
  Toderici}, \bibinfo{person}{Susanna Ricco}, \bibinfo{person}{Rahul
  Sukthankar}, \bibinfo{person}{Cordelia Schmid}, {and}
  \bibinfo{person}{Jitendra Malik}.} \bibinfo{year}{2018}\natexlab{}.
\newblock \showarticletitle{{AVA: A Video Dataset of Spatio-temporally
  Localized Atomic Visual Actions}}. In \bibinfo{booktitle}{\emph{Conference on
  Computer Vision and Pattern Recognition (CVPR)}}.
\newblock


\bibitem[\protect\citeauthoryear{He, Zhang, Ren, and Sun}{He
  et~al\mbox{.}}{2015}]%
        {he2015superhumanimagenet}
\bibfield{author}{\bibinfo{person}{Kaiming He}, \bibinfo{person}{Xiangyu
  Zhang}, \bibinfo{person}{Shaoqing Ren}, {and} \bibinfo{person}{Jian Sun}.}
  \bibinfo{year}{2015}\natexlab{}.
\newblock \showarticletitle{{Delving Deep into Rectifiers: Surpassing
  Human-Level Performance on ImageNet Classification}}. In
  \bibinfo{booktitle}{\emph{International Conference on Computer Vision
  (ICCV)}}.
\newblock


\bibitem[\protect\citeauthoryear{He, Zhang, Ren, and Sun}{He
  et~al\mbox{.}}{2016}]%
        {he2016resnet}
\bibfield{author}{\bibinfo{person}{Kaiming He}, \bibinfo{person}{Xiangyu
  Zhang}, \bibinfo{person}{Shaoqing Ren}, {and} \bibinfo{person}{Jian Sun}.}
  \bibinfo{year}{2016}\natexlab{}.
\newblock \showarticletitle{{Deep Residual Learning for Image Recognition}}. In
  \bibinfo{booktitle}{\emph{Conference on Computer Vision and Pattern
  Recognition (CVPR)}}.
\newblock


\bibitem[\protect\citeauthoryear{Hong, Han, Choi, Seo, Kim, and Chang}{Hong
  et~al\mbox{.}}{2021}]%
        {hong2021disentangling}
\bibfield{author}{\bibinfo{person}{Youngkyu Hong}, \bibinfo{person}{Seungju
  Han}, \bibinfo{person}{Kwanghee Choi}, \bibinfo{person}{Seokjun Seo},
  \bibinfo{person}{Beomsu Kim}, {and} \bibinfo{person}{Buru Chang}.}
  \bibinfo{year}{2021}\natexlab{}.
\newblock \showarticletitle{Disentangling Label Distribution for Long-tailed
  Visual Recognition}. In \bibinfo{booktitle}{\emph{CVPR}}.
\newblock


\bibitem[\protect\citeauthoryear{Iscen, Araujo, Gong, and Schmid}{Iscen
  et~al\mbox{.}}{2021}]%
        {iscen2021cbd}
\bibfield{author}{\bibinfo{person}{Ahmet Iscen}, \bibinfo{person}{André
  Araujo}, \bibinfo{person}{Boqing Gong}, {and} \bibinfo{person}{Cordelia
  Schmid}.} \bibinfo{year}{2021}\natexlab{}.
\newblock \showarticletitle{{Class-Balanced Distillation for Long-Tailed Visual
  Recognition}}. In \bibinfo{booktitle}{\emph{arXiv:2104.05279}}.
\newblock


\bibitem[\protect\citeauthoryear{Jamal, Brown, Yang, Wang, and Gong}{Jamal
  et~al\mbox{.}}{2020}]%
        {jamal2020domainadaptation}
\bibfield{author}{\bibinfo{person}{Muhammad~Abdullah Jamal},
  \bibinfo{person}{Matthew Brown}, \bibinfo{person}{Ming-Hsuan Yang},
  \bibinfo{person}{Liqiang Wang}, {and} \bibinfo{person}{Boqing Gong}.}
  \bibinfo{year}{2020}\natexlab{}.
\newblock \showarticletitle{Rethinking Class-Balanced Methods for Long-Tailed
  Visual Recognition from a Domain Adaptation Perspective}. In
  \bibinfo{booktitle}{\emph{Conference on Computer Vision and Pattern
  Recognition (CVPR)}}.
\newblock


\bibitem[\protect\citeauthoryear{Japkowicz}{Japkowicz}{2000}]%
        {japkowicz2000sampling}
\bibfield{author}{\bibinfo{person}{Nathalie Japkowicz}.}
  \bibinfo{year}{2000}\natexlab{}.
\newblock \showarticletitle{{The Class Imbalance Problem: Significance and
  Strategies}}. In \bibinfo{booktitle}{\emph{International Conference on
  Artificial Intelligence (IC-AI)}}.
\newblock


\bibitem[\protect\citeauthoryear{Kang, Xie, Rohrbach, Yan, Gordo, Feng, and
  Kalantidis}{Kang et~al\mbox{.}}{2020}]%
        {kang2019decoupling}
\bibfield{author}{\bibinfo{person}{Bingyi Kang}, \bibinfo{person}{Saining Xie},
  \bibinfo{person}{Marcus Rohrbach}, \bibinfo{person}{Zhicheng Yan},
  \bibinfo{person}{Albert Gordo}, \bibinfo{person}{Jiashi Feng}, {and}
  \bibinfo{person}{Yannis Kalantidis}.} \bibinfo{year}{2020}\natexlab{}.
\newblock \showarticletitle{Decoupling representation and classifier for
  long-tailed recognition}. In \bibinfo{booktitle}{\emph{International
  Conference on Learning Representations (ICLR)}}.
\newblock


\bibitem[\protect\citeauthoryear{Kim, Jeong, and Shin}{Kim
  et~al\mbox{.}}{2020}]%
        {kim2020m2m}
\bibfield{author}{\bibinfo{person}{Jaehyung Kim}, \bibinfo{person}{Jongheon
  Jeong}, {and} \bibinfo{person}{Jinwoo Shin}.}
  \bibinfo{year}{2020}\natexlab{}.
\newblock \showarticletitle{M2m: Imbalanced Classification via Major-to-minor
  Translation}. In \bibinfo{booktitle}{\emph{Conference on Computer Vision and
  Pattern Recognition (CVPR)}}.
\newblock


\bibitem[\protect\citeauthoryear{Law, Liao, Snell, and Zemel}{Law
  et~al\mbox{.}}{2019}]%
        {pmlr-v97-law19a}
\bibfield{author}{\bibinfo{person}{Marc Law}, \bibinfo{person}{Renjie Liao},
  \bibinfo{person}{Jake Snell}, {and} \bibinfo{person}{Richard Zemel}.}
  \bibinfo{year}{2019}\natexlab{}.
\newblock \showarticletitle{Lorentzian Distance Learning for Hyperbolic
  Representations}. In \bibinfo{booktitle}{\emph{International Conference on
  Machine Learning (ICML)}}.
\newblock


\bibitem[\protect\citeauthoryear{Law and Stam}{Law and Stam}{2020}]%
        {NEURIPS2020_123b7f02}
\bibfield{author}{\bibinfo{person}{Marc Law} {and} \bibinfo{person}{Jos Stam}.}
  \bibinfo{year}{2020}\natexlab{}.
\newblock \showarticletitle{Ultrahyperbolic Representation Learning}. In
  \bibinfo{booktitle}{\emph{Advances in Neural Information Processing Systems
  (NeurIPS)}}.
\newblock


\bibitem[\protect\citeauthoryear{Lin, Goyal, Girshick, He, and Doll{\'a}r}{Lin
  et~al\mbox{.}}{2017}]%
        {lin2017focal}
\bibfield{author}{\bibinfo{person}{Tsung-Yi Lin}, \bibinfo{person}{Priya
  Goyal}, \bibinfo{person}{Ross Girshick}, \bibinfo{person}{Kaiming He}, {and}
  \bibinfo{person}{Piotr Doll{\'a}r}.} \bibinfo{year}{2017}\natexlab{}.
\newblock \showarticletitle{Focal loss for dense object detection}. In
  \bibinfo{booktitle}{\emph{International Conference on Computer Vision
  (ICCV)}}.
\newblock


\bibitem[\protect\citeauthoryear{Liu, Sun, Han, Dou, and Li}{Liu
  et~al\mbox{.}}{2020}]%
        {liu2020headtailvariance}
\bibfield{author}{\bibinfo{person}{Jialun Liu}, \bibinfo{person}{Yifan Sun},
  \bibinfo{person}{Chuchu Han}, \bibinfo{person}{Zhaopeng Dou}, {and}
  \bibinfo{person}{Wenhui Li}.} \bibinfo{year}{2020}\natexlab{}.
\newblock \showarticletitle{Deep Representation Learning on Long-tailed Data: A
  Learnable Embedding Augmentation Perspective}. In
  \bibinfo{booktitle}{\emph{Conference on Computer Vision and Pattern
  Recognition (CVPR)}}.
\newblock


\bibitem[\protect\citeauthoryear{Liu, Miao, Zhan, Wang, Gong, and Yu}{Liu
  et~al\mbox{.}}{2019}]%
        {liu2019oltr}
\bibfield{author}{\bibinfo{person}{Ziwei Liu}, \bibinfo{person}{Zhongqi Miao},
  \bibinfo{person}{Xiaohang Zhan}, \bibinfo{person}{Jiayun Wang},
  \bibinfo{person}{Boqing Gong}, {and} \bibinfo{person}{Stella~X. Yu}.}
  \bibinfo{year}{2019}\natexlab{}.
\newblock \showarticletitle{{Large-Scale Long-Tailed Recognition in an Open
  World}}. In \bibinfo{booktitle}{\emph{Conference on Computer Vision and
  Pattern Recognition (CVPR)}}.
\newblock


\bibitem[\protect\citeauthoryear{Loshchilov and Hutter}{Loshchilov and
  Hutter}{2017}]%
        {coslr2016sgdr}
\bibfield{author}{\bibinfo{person}{Ilya Loshchilov} {and}
  \bibinfo{person}{Frank Hutter}.} \bibinfo{year}{2017}\natexlab{}.
\newblock \showarticletitle{{SGDR}: Stochastic gradient descent with warm
  restarts}. In \bibinfo{booktitle}{\emph{International Conference on Learning
  Representations (ICLR)}}.
\newblock


\bibitem[\protect\citeauthoryear{Luo, Zhan, Xue, Wang, Ren, and Yang}{Luo
  et~al\mbox{.}}{2018}]%
        {luo2018cosinece}
\bibfield{author}{\bibinfo{person}{Chunjie Luo}, \bibinfo{person}{Jianfeng
  Zhan}, \bibinfo{person}{Xiaohe Xue}, \bibinfo{person}{Lei Wang},
  \bibinfo{person}{Rui Ren}, {and} \bibinfo{person}{Qiang Yang}.}
  \bibinfo{year}{2018}\natexlab{}.
\newblock \showarticletitle{{Cosine normalization: Using cosine similarity
  instead of dot product in neural networks}}. In
  \bibinfo{booktitle}{\emph{International Conference on Artificial Neural
  Networks (ICANN)}}.
\newblock


\bibitem[\protect\citeauthoryear{Mangla, Singh, Sinha, Kumari, K, and
  Balasubramanian}{Mangla et~al\mbox{.}}{2020}]%
        {mangla2020manifoldfsl}
\bibfield{author}{\bibinfo{person}{Puneet Mangla}, \bibinfo{person}{Mayank
  Singh}, \bibinfo{person}{Abhishek Sinha}, \bibinfo{person}{Nupur Kumari},
  \bibinfo{person}{Balaji K}, {and} \bibinfo{person}{V~N Balasubramanian}.}
  \bibinfo{year}{2020}\natexlab{}.
\newblock \showarticletitle{{Charting the Right Manifold: Manifold Mixup for
  Few-shot Learning}}. In \bibinfo{booktitle}{\emph{Winter Conference on
  Applications of Computer Vision (WACV)}}.
\newblock


\bibitem[\protect\citeauthoryear{Menon, Jayasumana, Rawat, Jain, Veit, and
  Kumar}{Menon et~al\mbox{.}}{2021}]%
        {menon2021logitadjustment}
\bibfield{author}{\bibinfo{person}{Aditya~Krishna Menon},
  \bibinfo{person}{Sadeep Jayasumana}, \bibinfo{person}{Ankit~Singh Rawat},
  \bibinfo{person}{Himanshu Jain}, \bibinfo{person}{Andreas Veit}, {and}
  \bibinfo{person}{Sanjiv Kumar}.} \bibinfo{year}{2021}\natexlab{}.
\newblock \showarticletitle{{Long-tail Learning via Logit Adjustment}}. In
  \bibinfo{booktitle}{\emph{International Conference on Learning
  Representations (ICLR)}}.
\newblock


\bibitem[\protect\citeauthoryear{Miech, Zhukov, Alayrac, Tapaswi, Laptev, and
  Sivic}{Miech et~al\mbox{.}}{2019}]%
        {miech2019howto100m}
\bibfield{author}{\bibinfo{person}{Antoine Miech}, \bibinfo{person}{Dimitri
  Zhukov}, \bibinfo{person}{Jean-Baptiste Alayrac}, \bibinfo{person}{Makarand
  Tapaswi}, \bibinfo{person}{Ivan Laptev}, {and} \bibinfo{person}{Josef
  Sivic}.} \bibinfo{year}{2019}\natexlab{}.
\newblock \showarticletitle{{HowTo100M: Learning a Text-Video Embedding by
  Watching Hundred Million Narrated Video Clips}}. In
  \bibinfo{booktitle}{\emph{International Conference on Computer Vision
  (ICCV)}}.
\newblock


\bibitem[\protect\citeauthoryear{Min, Wang, Liu, Luo, Kang, Wei, Wei, and
  Jiang}{Min et~al\mbox{.}}{2021}]%
        {finefoodsworkshop}
\bibfield{author}{\bibinfo{person}{Weiqing Min}, \bibinfo{person}{Zhiling
  Wang}, \bibinfo{person}{Yuxin Liu}, \bibinfo{person}{Mengjiang Luo},
  \bibinfo{person}{Liping Kang}, \bibinfo{person}{Xiaoming Wei},
  \bibinfo{person}{Xiaolin Wei}, {and} \bibinfo{person}{Shuqiang Jiang}.}
  \bibinfo{year}{2021}\natexlab{}.
\newblock \showarticletitle{Large Scale Visual Food Recognition}.
\newblock \bibinfo{journal}{\emph{arXiv:2103.16107}} (\bibinfo{year}{2021}).
\newblock


\bibitem[\protect\citeauthoryear{Parkhi, Vedaldi, and Zisserman}{Parkhi
  et~al\mbox{.}}{2015}]%
        {parkhi2015vggface}
\bibfield{author}{\bibinfo{person}{Omkar~M. Parkhi}, \bibinfo{person}{Andrea
  Vedaldi}, {and} \bibinfo{person}{Andrew Zisserman}.}
  \bibinfo{year}{2015}\natexlab{}.
\newblock \showarticletitle{{Deep Face Recognition}}. In
  \bibinfo{booktitle}{\emph{British Machine Vision Conference (BMVC)}}.
\newblock


\bibitem[\protect\citeauthoryear{Prakash, Debnath, Lafleche, Cameracci, State,
  Birchfield, and Law}{Prakash et~al\mbox{.}}{2021}]%
        {sim2sg}
\bibfield{author}{\bibinfo{person}{Aayush Prakash}, \bibinfo{person}{Shoubhik
  Debnath}, \bibinfo{person}{Jean-Francois Lafleche}, \bibinfo{person}{Eric
  Cameracci}, \bibinfo{person}{Gavriel State}, \bibinfo{person}{Stan
  Birchfield}, {and} \bibinfo{person}{Marc~T. Law}.}
  \bibinfo{year}{2021}\natexlab{}.
\newblock \showarticletitle{{Self-Supervised Real-to-Sim Scene Generation}}. In
  \bibinfo{booktitle}{\emph{International Conference on Computer Vision
  (ICCV)}}.
\newblock


\bibitem[\protect\citeauthoryear{Ren, Zeng, Yang, and Urtasun}{Ren
  et~al\mbox{.}}{2018}]%
        {ren2018metareweight}
\bibfield{author}{\bibinfo{person}{Mengye Ren}, \bibinfo{person}{Wenyuan Zeng},
  \bibinfo{person}{Bin Yang}, {and} \bibinfo{person}{Raquel Urtasun}.}
  \bibinfo{year}{2018}\natexlab{}.
\newblock \showarticletitle{{Learning to Reweight Examples for Robust Deep
  Learning}}. In \bibinfo{booktitle}{\emph{International Conference on Machine
  Learning (ICML)}}.
\newblock


\bibitem[\protect\citeauthoryear{Russakovsky, Deng, Su, Krause, Satheesh, Ma,
  Huang, Karpathy, Khosla, Bernstein, Berg, and Fei-Fei}{Russakovsky
  et~al\mbox{.}}{2015}]%
        {russakovsky2015imagenet}
\bibfield{author}{\bibinfo{person}{Olga Russakovsky}, \bibinfo{person}{Jia
  Deng}, \bibinfo{person}{Hao Su}, \bibinfo{person}{Jonathan Krause},
  \bibinfo{person}{Sanjeev Satheesh}, \bibinfo{person}{Sean Ma},
  \bibinfo{person}{Zhiheng Huang}, \bibinfo{person}{Andrej Karpathy},
  \bibinfo{person}{Aditya Khosla}, \bibinfo{person}{Michael Bernstein},
  \bibinfo{person}{Alexander~C. Berg}, {and} \bibinfo{person}{Li Fei-Fei}.}
  \bibinfo{year}{2015}\natexlab{}.
\newblock \showarticletitle{{ImageNet Large Scale Visual Recognition
  Challenge}}.
\newblock \bibinfo{journal}{\emph{International Journal of Computer Vision
  (IJCV)}}  \bibinfo{volume}{115} (\bibinfo{year}{2015}),
  \bibinfo{pages}{211--252}.
\newblock


\bibitem[\protect\citeauthoryear{Sharma, Ding, Goodman, and Soricut}{Sharma
  et~al\mbox{.}}{2018}]%
        {sharma2018conceptual}
\bibfield{author}{\bibinfo{person}{Piyush Sharma}, \bibinfo{person}{Nan Ding},
  \bibinfo{person}{Sebastian Goodman}, {and} \bibinfo{person}{Radu Soricut}.}
  \bibinfo{year}{2018}\natexlab{}.
\newblock \showarticletitle{{Conceptual Captions: A Cleaned, Hypernymed, Image
  Alt-text Dataset for Automatic Image Captioning}}. In
  \bibinfo{booktitle}{\emph{Proceedings of ACL}}.
\newblock


\bibitem[\protect\citeauthoryear{Shu, Xie, Yi, Zhao, Zhou, Xu, and Meng}{Shu
  et~al\mbox{.}}{2019}]%
        {shu2019metaweightnet}
\bibfield{author}{\bibinfo{person}{Jun Shu}, \bibinfo{person}{Qi Xie},
  \bibinfo{person}{Lixuan Yi}, \bibinfo{person}{Qian Zhao},
  \bibinfo{person}{Sanping Zhou}, \bibinfo{person}{Zongben Xu}, {and}
  \bibinfo{person}{Deyu Meng}.} \bibinfo{year}{2019}\natexlab{}.
\newblock \showarticletitle{Meta-weight-net: Learning an explicit mapping for
  sample weighting}. In \bibinfo{booktitle}{\emph{Advances in Neural
  Information Processing Systems (NeurIPS)}}.
\newblock


\bibitem[\protect\citeauthoryear{Sun, Shrivastava, Singh, and Gupta}{Sun
  et~al\mbox{.}}{2017}]%
        {sun2017jft}
\bibfield{author}{\bibinfo{person}{Chen Sun}, \bibinfo{person}{Abhinav
  Shrivastava}, \bibinfo{person}{Saurabh Singh}, {and} \bibinfo{person}{Abhinav
  Gupta}.} \bibinfo{year}{2017}\natexlab{}.
\newblock \showarticletitle{{Revisiting Unreasonable Effectiveness of Data in
  Deep Learning Era}}. In \bibinfo{booktitle}{\emph{International Conference on
  Computer Vision (ICCV)}}.
\newblock


\bibitem[\protect\citeauthoryear{Tang, Huang, and Zhang}{Tang
  et~al\mbox{.}}{2020}]%
        {tang2020gbm}
\bibfield{author}{\bibinfo{person}{Kaihua Tang}, \bibinfo{person}{Jianqiang
  Huang}, {and} \bibinfo{person}{Hanwang Zhang}.}
  \bibinfo{year}{2020}\natexlab{}.
\newblock \showarticletitle{{Long-Tailed Classification by Keeping the Good and
  Removing the Bad Momentum Causal Effect}}. In
  \bibinfo{booktitle}{\emph{Advances in Neural Information Processing Systems
  (NeurIPS)}}.
\newblock


\bibitem[\protect\citeauthoryear{Tsipras, Santurkar, Engstrom, Ilyas, and
  Madry}{Tsipras et~al\mbox{.}}{2020}]%
        {tsipras2020imagenet}
\bibfield{author}{\bibinfo{person}{Dimitris Tsipras}, \bibinfo{person}{Shibani
  Santurkar}, \bibinfo{person}{Logan Engstrom}, \bibinfo{person}{Andrew Ilyas},
  {and} \bibinfo{person}{Aleksander Madry}.} \bibinfo{year}{2020}\natexlab{}.
\newblock \showarticletitle{{From ImageNet to Image Classification:
  Contextualizing Progress on Benchmarks}}. In
  \bibinfo{booktitle}{\emph{International Conference on Machine Learning
  (ICML)}}.
\newblock


\bibitem[\protect\citeauthoryear{Tukey}{Tukey}{1977}]%
        {tukey1977eda}
\bibfield{author}{\bibinfo{person}{John~W Tukey}.}
  \bibinfo{year}{1977}\natexlab{}.
\newblock \bibinfo{booktitle}{\emph{{Exploratory Data Analysis}}}.
\newblock \bibinfo{publisher}{Addison-Wesley}.
\newblock


\bibitem[\protect\citeauthoryear{van~der Maaten and Hinton}{van~der Maaten and
  Hinton}{2008}]%
        {maaten2008tsne}
\bibfield{author}{\bibinfo{person}{Laurens van~der Maaten} {and}
  \bibinfo{person}{Geoffrey~E. Hinton}.} \bibinfo{year}{2008}\natexlab{}.
\newblock \showarticletitle{{Visualizing Data Using t-SNE}}.
\newblock \bibinfo{journal}{\emph{Journal of Machine Learning Recognition
  (JMLR)}}  \bibinfo{volume}{9} (\bibinfo{year}{2008}),
  \bibinfo{pages}{2579--2605}.
\newblock


\bibitem[\protect\citeauthoryear{van Horn, Aodha, Song, Cui, Sun, Shepard,
  Adam, Perona, and Belongie}{van Horn et~al\mbox{.}}{2018}]%
        {vanhorn2018inaturalist}
\bibfield{author}{\bibinfo{person}{Grant van Horn}, \bibinfo{person}{Oisin~Mac
  Aodha}, \bibinfo{person}{Yang Song}, \bibinfo{person}{Yin Cui},
  \bibinfo{person}{Chen Sun}, \bibinfo{person}{Alex Shepard},
  \bibinfo{person}{Hartwig Adam}, \bibinfo{person}{Pietro Perona}, {and}
  \bibinfo{person}{Serge Belongie}.} \bibinfo{year}{2018}\natexlab{}.
\newblock \showarticletitle{{The iNaturalist Species Classification and
  Detection Dataset}}. In \bibinfo{booktitle}{\emph{Conference on Computer
  Vision and Pattern Recognition (CVPR)}}.
\newblock


\bibitem[\protect\citeauthoryear{Verma, Lamb, Beckham, Najafi, Mitliagkas,
  Courville, Lopez-Paz, and Bengio}{Verma et~al\mbox{.}}{2019}]%
        {verma2019manifoldmixup}
\bibfield{author}{\bibinfo{person}{Vikas Verma}, \bibinfo{person}{Alex Lamb},
  \bibinfo{person}{Christopher Beckham}, \bibinfo{person}{Amir Najafi},
  \bibinfo{person}{Ioannis Mitliagkas}, \bibinfo{person}{Aaron Courville},
  \bibinfo{person}{David Lopez-Paz}, {and} \bibinfo{person}{Yoshua Bengio}.}
  \bibinfo{year}{2019}\natexlab{}.
\newblock \showarticletitle{{Manifold Mixup: Better Representations by
  Interpolating Hidden States}}. In \bibinfo{booktitle}{\emph{International
  Conference on Machine Learning (ICML)}}.
\newblock


\bibitem[\protect\citeauthoryear{Vicol, Tapaswi, Castrejon, and Fidler}{Vicol
  et~al\mbox{.}}{2018}]%
        {vicol2018moviegraphs}
\bibfield{author}{\bibinfo{person}{Paul Vicol}, \bibinfo{person}{Makarand
  Tapaswi}, \bibinfo{person}{Lluis Castrejon}, {and} \bibinfo{person}{Sanja
  Fidler}.} \bibinfo{year}{2018}\natexlab{}.
\newblock \showarticletitle{{MovieGraphs: Towards Understanding Human-Centric
  Situations from Videos}}. In \bibinfo{booktitle}{\emph{Conference on Computer
  Vision and Pattern Recognition (CVPR)}}.
\newblock


\bibitem[\protect\citeauthoryear{Vinyals, Blundell, Lillicrap, Wierstra,
  et~al\mbox{.}}{Vinyals et~al\mbox{.}}{2016}]%
        {vinyals2016matching}
\bibfield{author}{\bibinfo{person}{Oriol Vinyals}, \bibinfo{person}{Charles
  Blundell}, \bibinfo{person}{Timothy Lillicrap}, \bibinfo{person}{Daan
  Wierstra}, {et~al\mbox{.}}} \bibinfo{year}{2016}\natexlab{}.
\newblock \showarticletitle{Matching networks for one shot learning}. In
  \bibinfo{booktitle}{\emph{Advances in Neural Information Processing Systems
  (NeurIPS)}}.
\newblock


\bibitem[\protect\citeauthoryear{Wang, Singh, Michael, Hill, Levy, and
  Bowman}{Wang et~al\mbox{.}}{2019a}]%
        {wang2019glue}
\bibfield{author}{\bibinfo{person}{Alex Wang}, \bibinfo{person}{Amanpreet
  Singh}, \bibinfo{person}{Julian Michael}, \bibinfo{person}{Felix Hill},
  \bibinfo{person}{Omer Levy}, {and} \bibinfo{person}{Samuel~R. Bowman}.}
  \bibinfo{year}{2019}\natexlab{a}.
\newblock \showarticletitle{{GLUE: A Multi-task Benchmark and Analysis Platform
  for Natural Language Understanding}}. In
  \bibinfo{booktitle}{\emph{International Conference on Learning
  Representations (ICLR)}}.
\newblock


\bibitem[\protect\citeauthoryear{Wang, Wang, Law, Rudzicz, and Brudno}{Wang
  et~al\mbox{.}}{2019b}]%
        {8683393}
\bibfield{author}{\bibinfo{person}{Jixuan Wang}, \bibinfo{person}{Kuan-Chieh
  Wang}, \bibinfo{person}{Marc~T. Law}, \bibinfo{person}{Frank Rudzicz}, {and}
  \bibinfo{person}{Michael Brudno}.} \bibinfo{year}{2019}\natexlab{b}.
\newblock \showarticletitle{Centroid-based Deep Metric Learning for Speaker
  Recognition}. In \bibinfo{booktitle}{\emph{IEEE International Conference on
  Acoustics, Speech and Signal Processing (ICASSP)}}.
\newblock


\bibitem[\protect\citeauthoryear{Wang, Lian, Miao, Liu, and Yu}{Wang
  et~al\mbox{.}}{2021}]%
        {wang2020long}
\bibfield{author}{\bibinfo{person}{Xudong Wang}, \bibinfo{person}{Long Lian},
  \bibinfo{person}{Zhongqi Miao}, \bibinfo{person}{Ziwei Liu}, {and}
  \bibinfo{person}{Stella~X Yu}.} \bibinfo{year}{2021}\natexlab{}.
\newblock \showarticletitle{Long-tailed recognition by routing diverse
  distribution-aware experts}. In \bibinfo{booktitle}{\emph{International
  Conference on Learning Representations (ICLR)}}.
\newblock


\bibitem[\protect\citeauthoryear{Wang, Yao, Kwok, and Ni}{Wang
  et~al\mbox{.}}{2019c}]%
        {wang2019fslsurvey}
\bibfield{author}{\bibinfo{person}{Yaqing Wang}, \bibinfo{person}{Quanming
  Yao}, \bibinfo{person}{James Kwok}, {and} \bibinfo{person}{Lionel~M. Ni}.}
  \bibinfo{year}{2019}\natexlab{c}.
\newblock \showarticletitle{{Generalizing from a Few Examples: A Survey on
  Few-Shot Learning}}. In \bibinfo{booktitle}{\emph{arXiv:1904.05046}}.
\newblock


\bibitem[\protect\citeauthoryear{Xiang, Ding, and Han}{Xiang
  et~al\mbox{.}}{2020}]%
        {xiang2020lfme}
\bibfield{author}{\bibinfo{person}{Liuyu Xiang}, \bibinfo{person}{Guiguang
  Ding}, {and} \bibinfo{person}{Jungong Han}.} \bibinfo{year}{2020}\natexlab{}.
\newblock \showarticletitle{{Learning From Multiple Experts: Self-paced
  Knowledge Distillation for Long-tailed Classification}}. In
  \bibinfo{booktitle}{\emph{European Conference on Computer Vision (ECCV)}}.
\newblock


\bibitem[\protect\citeauthoryear{Yang, Liu, and Xu}{Yang et~al\mbox{.}}{2021}]%
        {yang2021freelunch}
\bibfield{author}{\bibinfo{person}{Shuo Yang}, \bibinfo{person}{Lu Liu}, {and}
  \bibinfo{person}{Min Xu}.} \bibinfo{year}{2021}\natexlab{}.
\newblock \showarticletitle{{Free Lunch for Few-Shot Learning: Distribution
  Calibration}}. In \bibinfo{booktitle}{\emph{International Conference on
  Learning Representations (ICLR)}}.
\newblock


\bibitem[\protect\citeauthoryear{Zhang, Cisse, Dauphin, and Lopez-Paz}{Zhang
  et~al\mbox{.}}{2017}]%
        {zhang2017mixup}
\bibfield{author}{\bibinfo{person}{Hongyi Zhang}, \bibinfo{person}{Moustapha
  Cisse}, \bibinfo{person}{Yann~N. Dauphin}, {and} \bibinfo{person}{David
  Lopez-Paz}.} \bibinfo{year}{2017}\natexlab{}.
\newblock \showarticletitle{{mixup: Beyond Empirical Risk Minimization}}. In
  \bibinfo{booktitle}{\emph{International Conference on Learning
  Representations (ICLR)}}.
\newblock


\bibitem[\protect\citeauthoryear{Zhong, Cui, Liu, and Jia}{Zhong
  et~al\mbox{.}}{2021}]%
        {zhong2021improving}
\bibfield{author}{\bibinfo{person}{Zhisheng Zhong}, \bibinfo{person}{Jiequan
  Cui}, \bibinfo{person}{Shu Liu}, {and} \bibinfo{person}{Jiaya Jia}.}
  \bibinfo{year}{2021}\natexlab{}.
\newblock \showarticletitle{Improving Calibration for Long-Tailed Recognition}.
  In \bibinfo{booktitle}{\emph{CVPR}}.
\newblock


\bibitem[\protect\citeauthoryear{Zhou, Cui, Wei, and Chen}{Zhou
  et~al\mbox{.}}{2020}]%
        {zhou2020bbn}
\bibfield{author}{\bibinfo{person}{Boyan Zhou}, \bibinfo{person}{Quan Cui},
  \bibinfo{person}{Xiu-Shen Wei}, {and} \bibinfo{person}{Zhao-Min Chen}.}
  \bibinfo{year}{2020}\natexlab{}.
\newblock \showarticletitle{BBN: Bilateral-Branch Network with Cumulative
  Learning for Long-Tailed Visual Recognition}. In
  \bibinfo{booktitle}{\emph{Conference on Computer Vision and Pattern
  Recognition (CVPR)}}.
\newblock


\bibitem[\protect\citeauthoryear{Zipf}{Zipf}{1949}]%
        {zipf1949}
\bibfield{author}{\bibinfo{person}{G.~K. Zipf}.}
  \bibinfo{year}{1949}\natexlab{}.
\newblock \bibinfo{booktitle}{\emph{{Human behavior and the principle of least
  effort}}}.
\newblock \bibinfo{publisher}{Addison-Wesley Press}.
\newblock


\end{thebibliography}

\end{document}